\documentclass[10pt,twocolumn,letterpaper]{article}

\usepackage[pagenumbers]{cvpr} 

\usepackage{graphicx}
\usepackage{amsmath}
\usepackage{amssymb}
\usepackage{booktabs}
\usepackage{multirow}

\usepackage[pagebackref=true,breaklinks=true,colorlinks,bookmarks=false]{hyperref}

\usepackage[capitalize]{cleveref}
\crefname{section}{Sec.}{Secs.}
\Crefname{section}{Section}{Sections}
\Crefname{table}{Table}{Tables}
\crefname{table}{Tab.}{Tabs.}

\def\cvprPaperID{****} 
\def\confName{CVPR}
\def\confYear{2022}

\begin{document}

\title{Global Matching with Overlapping Attention for Optical Flow Estimation}


\author{Shiyu Zhao\textsuperscript{1,}\thanks{Correspondence
to: Shiyu Zhao (sz553@rutgers.edu).} \qquad 
Long Zhao\textsuperscript{2}  \qquad 
Zhixing Zhang\textsuperscript{1}  \qquad 
Enyu Zhou\textsuperscript{3} \qquad 
Dimitris Metaxas\textsuperscript{1}\\
\\
\textsuperscript{1}Rutgers University \qquad
\textsuperscript{2}Google Research\qquad
\textsuperscript{3}SenseTime Research\\
}

\maketitle

\begin{abstract}
   Optical flow estimation is a fundamental task in computer vision. Recent direct-regression methods using deep neural networks achieve remarkable performance improvement. However, they do not explicitly capture long-term motion correspondences and thus cannot handle large motions effectively.
   In this paper, inspired by the traditional matching-optimization methods where matching is introduced to handle large displacements before energy-based optimizations, we introduce a simple but effective global matching step before the direct regression and develop a learning-based matching-optimization framework, namely GMFlowNet. 
   In GMFlowNet, global matching is efficiently calculated by applying argmax on 4D cost volumes. 
   Additionally, to improve the matching quality, we propose patch-based overlapping attention to extract large context features. 
   Extensive experiments demonstrate that GMFlowNet outperforms RAFT, the most popular optimization-only method, by a large margin and achieves state-of-the-art performance on standard benchmarks. Thanks to the matching and overlapping attention, GMFlowNet obtains major improvements on the predictions for textureless regions and large motions. Our code is made publicly available at \url{https://github.com/xiaofeng94/GMFlowNet}.
\end{abstract}

\section{Introduction}
\label{sec:intro}

Optical flow estimation is a key computer vision task, which benefits various applications, including video interpolation \cite{jiang2018super}, deblurring \cite{yuan2020efficient}, video segmentation \cite{tsai2016video} and action recognition \cite{simonyan2014two}. 
Prevalent work in this area has been largely dominated by either matching-optimization or direct-regression methods.
Previous energy-based optimization methods \cite{horn1981determining,brox2004high,papenberg2006highly} usually fail to handle large displacements due to their inability to capture long-term motion correspondences. 
To remedy this, matching-optimization methods \cite{brox2010large,weinzaepfel2013deepflow,bailer2015flow} introduce a matching step before the optimization, which aims to find correspondences between pixels or patches across frames. 
However, their matching process depends on complicated hand-crafted features and is time-consuming and inaccurate.

\begin{figure}
  \centering
  \begin{subfigure}{1\linewidth}
    \centering
        \includegraphics[width=1\linewidth]{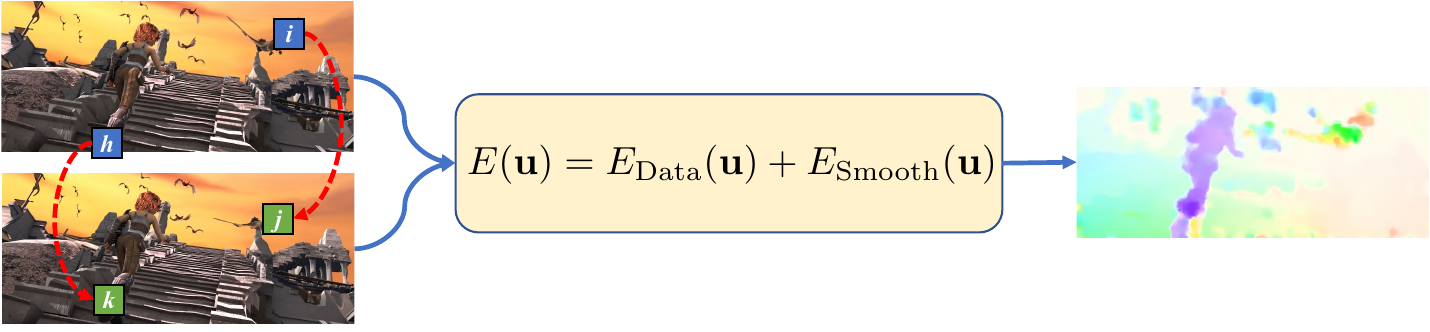}
    \caption{Traditional Matching-Optimization}
    \label{fig:matching}
  \end{subfigure}
  \begin{subfigure}{1\linewidth}
    \centering
        \includegraphics[width=1\linewidth]{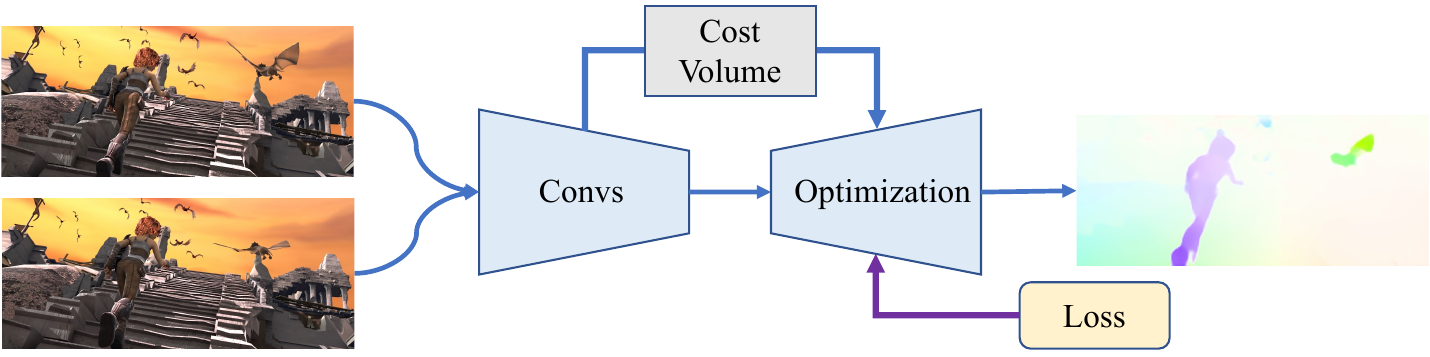}
    \caption{Learning-based Optimization}
    \label{fig:learning}
  \end{subfigure}
  \begin{subfigure}{1\linewidth}
    \centering
        \includegraphics[width=1\linewidth]{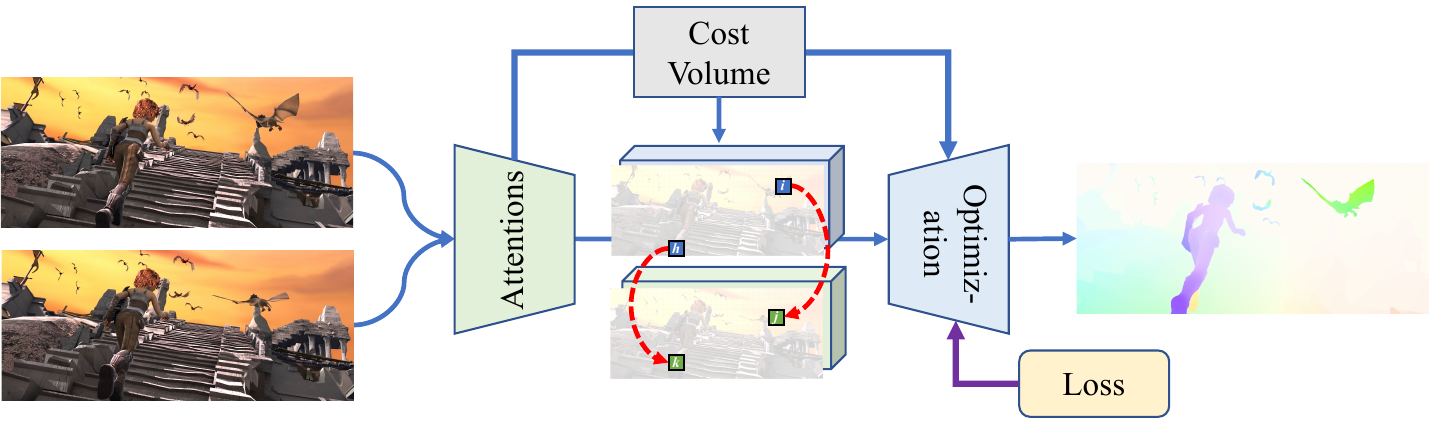}
    \caption{Learning-based Matching-Optimization}
    \label{fig:ours}
  \end{subfigure}
  \caption{\textbf{Main frameworks for optical flow estimation.} (a) Traditional matching-optimization methods first build a sparse matching to get a coarse flow and then exploit energy-based optimization to refine the flow. (b) Direct-regression methods mimic the energy-based optimization with learned parameters. They can be regarded as learning-based optimizations without matching. (c) Our framework introduces matching before the learning-based optimization and further improves the performance.
  }\label{fig:fig1}
\end{figure}

Recent direct-regression methods~\cite{sun2018pwc,hui2018liteflownet,teed2020raft,zhang2021separable} regard optical flow estimation as a regression task and achieve considerable improvements especially in predicting small changes in optical flow. These methods typically calculate 4D cost volumes representing the similarity between pixels and then directly regress flows from cost volumes by neural networks.
Similar to energy-based optimization, direct-regression methods cannot capture long-term motion correspondences in an explicit way and thus suffer from a performance drop in areas with large motions.

In this paper, we incorporate a matching step to explicitly handle large displacements for direct-regression methods, inspired by the improvement matching-optimization methods brought to energy-based optimization approaches.
Based on this idea, we develop a novel framework for optical flow estimation, namely Global Matching Flow Network (GMFlowNet), where global matching is introduced before the direct regression. 
Unlike traditional methods, GMFlowNet provides an efficient and accurate matching step.
For efficiency, we apply argmax to the typical 4D cost volume to build the global matching since it results in minor computational overhead. 
For accuracy, we propose a Patch-based OverLapping Attention  (POLA) block to extract large context features to diminish regional ambiguities in matching, e.g., repeated patterns and textureless regions. 
Specifically, POLA divides input feature maps into patches and attends each patch with itself and its neighboring patches. 
Since direct-regression methods mimic the traditional energy-based optimizations in a data-driven manner~\cite{teed2020raft}, they can be interpreted as learning-based optimizations. Thus, our method can be regarded as a learning-based matching-optimization framework. 
Fig.~\ref{fig:fig1} illustrates differences between previous related frameworks and ours.

We evaluate GMFlowNet on standard datasets for optical flow estimation.
Extensive experiments demonstrate that GMFlowNet significantly outperforms the most popular optimization-only model RAFT \cite{teed2020raft} and achieves state-of-the-art performance. 
As expected, GMFlowNet provides better flow estimations especially for large motion areas and textureless regions.
Besides, we thoroughly investigate our global matching and POLA, showing that they are both effective and efficient.

Our contributions are summarized as follows: 
1) We introduce a global matching step to explicitly handle large displacement optical flow estimations for direct-regression methods. With typical 4D cost volumes, our global matching is effective and efficient.
2) We propose a well-designed Patch-based OverLapping Attention (POLA) to address local ambiguities in matching and demonstrate its effectiveness via extensive experiments.
3) Following traditional matching-optimization frameworks, we propose a learning-based matching-optimization framework named GMFlowNet that achieves state of the art  performance on standard benchmarks.


\begin{figure*}
  \centering
  \includegraphics[width=1\linewidth]{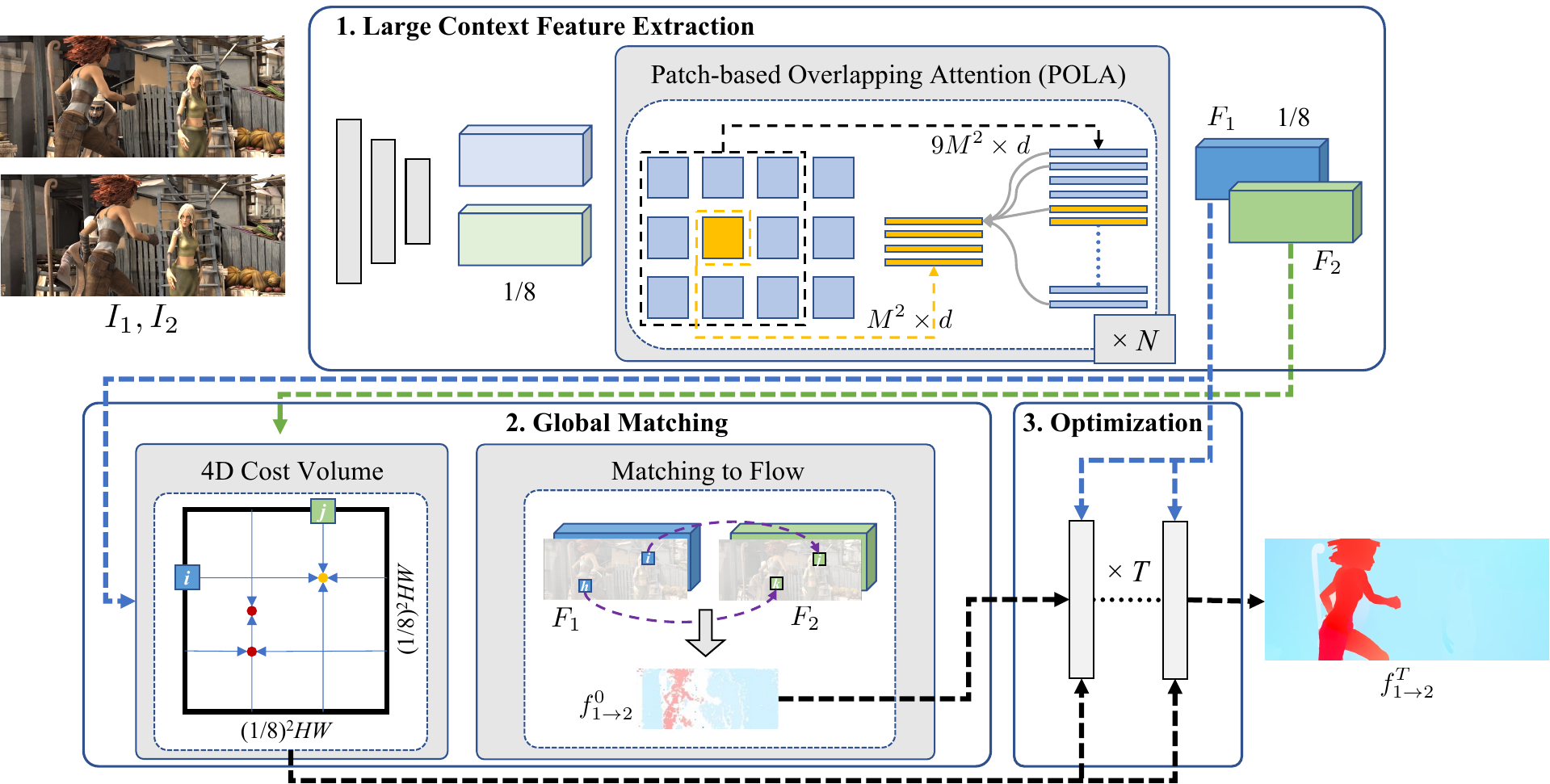}
  \caption{\textbf{Overview of GMFlowNet.} GMFlowNet has three components: {\bf 1)} The large context feature extraction module generates initial features from 3 convolutional layers and adopts the proposed POLA to extract large context information. $N$ refers to the number of attention blocks. {\bf 2)} The global matching module adopts large context features and constructs a 4D cost volume. Then, a global matching is built by applying argmax on the cost volume and refined by mutual matching. A coarse flow $f_{1\to 2}^0$ is generated from the matching. {\bf 3)} The optimization module takes $f_{1\to 2}^0$ as the initial state and updates the flow estimation iteratively. $T$ refers to the number of iterations. We employ the off-the-shelf optimization from RAFT \cite{teed2020raft}.
  } \label{fig:framework}
\end{figure*}

\section{Related Work}
\label{sec:related_work}

{\bf Optical flow as energy optimization.}
Previous methods formulated the optical flow as a continuous global energy function optimization problem~\cite{horn1981determining}. 
Black and Anandan~\cite{black1996cviu} introduced a robust estimation framework to address outliers caused by occlusions or significant brightness variations. Later research made further improvements by using better regularization terms~\cite{brox2004high, ranftl2014, zach2007pr} or additional robust optimization terms~\cite{black1996cviu, bruhn2002}. 
However, these approaches lack the ability to compute long-term dependencies and, thus, only work well for small displacements. 
To handle large displacements, later methods~\cite{Bruhn2005IJCV, brox2004high} introduced the coarse-to-fine strategy where large and small displacements are handled at different levels of an image pyramid.

However, coarse-to-fine approaches can neither handle small and fast-moving objects that disappear at coarse levels nor remedy mistakes made in the early stages. 
To address those issues, Brox and Malik~\cite{brox2010large} introduced feature matching to the energy-based optimization framework, which was further improved in later works~\cite{weinzaepfel2013deepflow, xu2011motion, revaud2015epicflow}. 
Following studies~\cite{Bai2016ExploitingSI, Bailer2017CNNBasedPM, chen2016full, hu2016efficient, Xu2017AccurateOF} widely adopted this approach. 
However, all these studies consider global matching highly time-consuming, so they only conducted local matching for computation efficiency, e.g., EpicFlow~\cite{revaud2015epicflow}. 
Contrary to previous methods, we calculate global matching efficiently by applying the argmax operator on widely adopted 4D cost volumes and achieve better performance.

{\bf Optical flow as network regression.}
More recently, the community has been motivated by the success of CNNs on high-level vision tasks~\cite{krizhevsky2012imagenet} to exploit learning-based solutions for optical flow estimation. 
Relevant studies ~\cite{dosovitskiy2015flownet, bar2020scopeflow, hur2019iterative, sun2018pwc, Zhao2020MaskFlownetAF, teed2020raft, hofinger2020improving, zhang2021separable} typically formulate optical flow estimation as regression instead of matching. 
In regression, cost volumes are the critical component that represents the similarity between pixels. 
For example, Sun et al.~\cite{sun2018pwc} designed a network using stacked image pyramids, feature warping, and cost volumes. 
Hofinger et al.~\cite{hofinger2020improving} employed a sampling-based strategy to improve the calculation of cost volumes. 
Teed and Deng~\cite{teed2020raft} built $4D$ cost volumes for all pairs of pixels. 
However, due to the high cost of memory and time, they did not aggregate the cost volumes to involve the global information. 
Separable Flow~\cite{zhang2021separable} proposed a separable cost volume module for efficient aggregations. 
In this work, 
we sidestep the high-cost global aggregation and leverage global information by constructing global matching using existing $4D$ cost volumes. 

{\bf Attention mechanism in vision.} 
This work extracts large context information for matching via leveraging recent advances in Vision Transformers~\cite{carion2020end, dosovitskiy2021an, liu2021swin}.
Methods leveraging Transformers' ability of modeling long-term dependencies have outperformed convolutional neural networks in various high-level computer vision tasks~\cite{dosovitskiy2021an, anonymous2022patches}. 
Inspired by these, Jiang et al.~\cite{jiang2021learning} introduced an attention-based module to resolve occlusions for optical flow estimation. 
Furthermore, LoFTR~\cite{sun2021loftr} adopted the self- and cross-attention to extract better descriptors for feature matching. 
Prevailing Vision Transformer architectures, e.g., Swin Transformer~\cite{liu2021swin}, conduct indirect inter-patch information exchange with shifted windows. 
We propose POLA to exchange information across patches directly.


\section{Approach}

We propose a novel framework GMFlowNet where a simple and effective global matching is introduced before the learning-based optimization.
Our GMFlowNet consists of three modules, namely, large context feature extraction, global matching, and learning-based optimization. Fig.~\ref{fig:framework} provides an overview of GMFlowNet, and each module is elaborated in the following sections.

\subsection{Large Context Feature Extraction}

Large context information is the key to handle matching in locally ambiguous locations, e.g. repeated patterns and textureless regions. GMFlowNet first employs 3 convolutional layers (3-Convs) to extract initial features and then adopts Transformer blocks to include long-term dependency information. 
Due to the large dimension of image features,
it's computationally prohibitive to apply vanilla self-attention \cite{vaswani2017attention} on whole feature maps.
To reduce the computation cost, we propose a well-designed local attention module POLA for optical flow estimation. 
In this section, we first describe attention in Tranformer and then we introduce POLA. In the end, we compare POLA with other feature extractors and discuss why ours is better for our task.

{\bf Attention in Transformer.} Given query vectors $Q \in \mathbb{R}^{N_q \times d}$, key vectors $K \in \mathbb{R}^{N_k \times d}$, and value vectors $V \in \mathbb{R}^{N_v \times d}$, where $d$ is the feature dimension, attention module attends $Q$ with $V$ by the similarity between $Q$ and $K$. Additionally, Ramachandran \emph{et al.} \cite{ramachandran2019stand} suggest a learned relative position bias $B \in \mathbb{R}^{N_q \times N_k}$ for better performance, and the attention is calculated as,
\begin{align}
    \text{Attention}(Q,K,V) = \text{softmax}(Q K^T/\sqrt{d} + B)\cdot V .
\end{align}
For more details about Transformers, please refer to \cite{vaswani2017attention}.

\begin{figure}
  \begin{subfigure}{0.6\linewidth}
    \centering
    \includegraphics[width=1\linewidth]{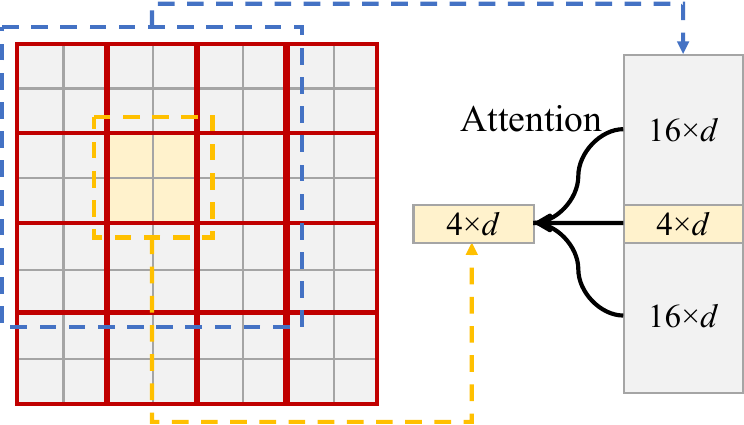}
    \caption{}
    \label{fig:overlap_attn}
  \end{subfigure}
  \hfill
  \begin{subfigure}{0.3\linewidth}
    \centering
    \includegraphics[width=1\linewidth]{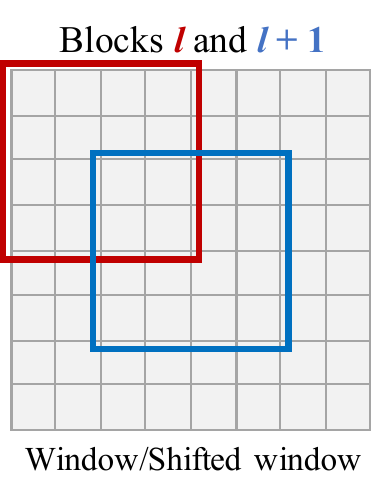}
    \caption{}
    \label{fig:shfit_win}
  \end{subfigure}
  \caption{{\bf Local attention.} (a) The proposed POLA.
  (b) Window partitions in Swin Transformer \cite{liu2021swin}. Red boxes highlight windows, and blue boxes highlight shifted windows. Two blocks are required to propagate information between windows.}
  \label{fig:attention}
\end{figure}

{\bf Patch-based overlapping attention.} 
Our POLA divides features into $M\times M$ non-overlapping patches and attends every patch with itself and its eight neighboring patches. Fig.~\ref{fig:overlap_attn} illustrates our POLA with $ M=2$. 
Following prior work \cite{vaswani2017attention,liu2021swin}, we adopt multi-head attentions in our attention block, as well. 
Given a patch vectorized as $P \in \mathbb{R}^{M^2 \times d}$ and its surrounding $3\times 3$ patches vectorized as $S \in \mathbb{R}^{9M^2 \times d}$, for the $i$-th head of our attention, we first project $P$ and $S$ into $d_k$ dimensions by learned linear projections and denote the projected results as $P_i$ and $S_i$, respectively. Then, we perform attention with $P_i$ and $S_i$ and get the output $h_i$. Finally, we concatenate $h_i$ from all heads as $H$ and project $H$ to $d$ dimensions as the final result $O \in \mathbb{R}^{M^2 \times d}$. Our multi-head patch-based overlapping attention can be formulated as,
\begin{align}
    h_i &= \text{Attention}(L^Q_i(P), L^K_i(S), L^V_i(S)). \nonumber \\
    H &= \text{Concat}([h_1,h_2,\dots,h_n])  \nonumber \\
    O &= L^O (H).
\end{align}
Here $n$ is the number of heads, $L^Q_i$, $L^K_i$, $L^V_i$ and $L^O$ are linear projection functions. In the experiments, we set $n = 8$ and $d_k = d/n$.

{\bf Why POLA is an improved attention method.}
Swin Transformer \cite{liu2021swin} provides a general local attention mechanism for vision tasks with windows and shifted windows as shown in Fig.~\ref{fig:shfit_win}.
However, the shifted window scheme requires two individual attention blocks to propagates inter-patch features, leading to information loss. Such loss is especially detrimental to matching because matching heavily depends on context information to reduce local ambiguities.
By contrast, our POLA involves inter-patch features within one block and propagates information directly with less information loss.
Moreover, POLA can be viewed as a generalization of per pixel overlapping attention that has been explored in \cite{ramachandran2019stand, hu2019local_all}. 
Compared with the per-pixel one, POLA enjoys at least three advantages: 1) consumes less memory, 2) can be efficiently implemented in existing deep learning platforms, and 3) arranges features by patch, which may provide better performance as suggested in recent research \cite{dosovitskiy2021an,liu2021swin,anonymous2022patches}.

\subsection{Global Matching}

We extract the context features $F_1$ and $F_2$ for the first input image $I_1$ and the second input image $I_2$, respectively. Then, a 4D cost volume is constructed on $F_1$ and $F_2$. After that, a global matching is computed from the cost volume and outputs a coarse flow $f_{1 \to 2}^0$ for $I_1$ and $I_2$, which is taken as the initial state of the later optimization.

{\bf 4D cost volume calculation.} We follow prior work \cite{teed2020raft,jiang2021learning} to construct the 4D cost volume on $1/8$ of the input resolution. The cost volume $C$ is calculated as,
\begin{align}
    C(i,j,u,v) = F_1(i,j) \cdot F_2(u,v),
\end{align}
where $(i,j)$ and $(u,v)$ refer to locations in $F_1$ and $F_2$.

{\bf Matching confidence calculation.} We adopt a dual-softmax operator \cite{Rocco18b} to convert the cost volume into matching confidence. This operator is efficient and enables the supervision of matching. In our case, the matching confidence $P_c$ is computed by,
\begin{align}
    P_c(i,j,u,v) = \text{softmax}(C(i,j, \cdot)) \odot \text{softmax}(C(\cdot, u, v)), 
\end{align}
where $C(i,j, \cdot)$ means all $(u,v)$ for given $(i,j)$. $C(\cdot, u, v)$ is similar. Pixel-wise production is denoted as $\odot$.

{\bf Matching selection and flow generation.} Based on $P_c$, we obtain the matching for $I_1$ at $(i,j)$ as
\begin{align}
    M_{1\to2}(i,j) = \mathop{\arg\max}_{u,v} P_c(i,j,u,v).
\end{align}
The matching for $I_2$, $M_{2\to1}(u,v)$, is attained similarly. Then, we pick robust matches that satisfy both $M_{1\to2}(i,j)$ and $M_{2\to1}(u,v)$ and define the matching set $M_c$ as,
\begin{align}
    M_c = \{(\hat{i},\hat{j})| (\hat{i},\hat{j}) = M_{2\to1}(M_{1\to2}(\hat{i},\hat{j}))\}.
\end{align}
The coarse flow is computed as,
\begin{align}
f_{1\to 2}^0 = \begin{cases}
    M_{1\to2}(i,j) - (i,j)& (i,j) \in M_c\\
    (0,0)& \text{Otherwise}
    \end{cases} .
\end{align}

\subsection{Optimization}
We use the off-the-shelf update operator from RAFT \cite{teed2020raft} as our optimization. This optimization predicts a delta flow and adds it to the current flow estimation. It iterates on such additions and outputs a series of flow predictions $\{f_{1\to 2}^1,f_{1\to 2}^2,\dots,f_{1\to 2}^T\}$, where $T$ is the total number of iterations and $f_{1\to 2}^T$ is used as the final prediction. 
We initialize the optimization with our coarse flow $f_{1\to 2}^0$ instead of the zero flow used in \cite{teed2020raft}. 
The optimization part in GMFlowNet is replaceable. We adopt RAFT's because it achieves the best performance. Any future optimization may be applied here for further improvements.

\subsection{Supervision}

{\bf Matching loss.} We round the ground truth optical flow $f_{1\to 2}^{gt}$ to the pixel level and collect the ground truth matching set $M_c^{gt}$. 
We consider regions as matched if they appear in both frames and set occlusion areas as unmatched. As the supervision in feature matching \cite{sun2021loftr}, we minimize the negative log-likelihood of $P_c$ in matched regions as,
\begin{align}
L_M = - \frac{1}{|M_c^{gt}|} \sum_{(\hat{i},\hat{j}) \in M_c^{gt}} \log P_c (\hat{i},\hat{j})
\end{align}

{\bf Optimization loss.} We follow RAFT \cite{teed2020raft} and supervise the optimization with $\ell_{1}$ distance between the predicted flow and $f_{gt}$. The optimization loss is defined as,
\begin{align}
L_O = \sum_{i=1}^{T} \gamma^{(i-T)}|| f_{1\to 2}^{gt} - f_{1\to 2}^i ||_1. 
\end{align}
The overall loss function of GMFlowNet is, 
\begin{align}
L = L_O + \lambda L_M 
\end{align}
where $\lambda$ balances different loss terms.


\section{Experiments}

This section elaborates on the experimental results to demonstrate the effectiveness of GMFlowNet. We show that GMFlowNet improves optical flow estimation when large motions and textureless regions are present based on both quantitative and qualitative evaluations. We also discuss the improvements in the results. 
An ablation study and an efficiency evaluation finalize the evaluation.

We implemented GMFlowNet in PyTorch \cite{paszke2019pytorch} and followed the training setting of RAFT\cite{teed2020raft}. 
We first train our model on FlyingChairs\cite{ilg2017flownet} (C) for 120k iterations (batch size of 10) and then finetune it on FlyingThings\cite{mayer2016large} (T) for 160k iterations (batch size of 6). 
After that, our model is further finetuned on a combination of data from FlyingThings (T), Sintel with both clean and final passes \cite{butler2012naturalistic} (S), KITTI\cite{menze2015object} (K), and/or HD1K\cite{kondermann2016hci} (H).
In the following sections, C+T refers to FlyingChairs and FlyingThings. C+T+S/K means C+T with either Sintel or KITTI. C+T+S+K+H refers to all training datasets.
We set the patch size to $M=7$ for POLA and the feature dimension to $d=256$.
When evaluating on Sintel, we improve the model by replacing 4 heads of our POLA blocks with 2 vertical and 2 horizontal axial-attention heads that are proposed by Wang \emph{et al.} \cite{wang2020axial}.

\subsection{Quantitative Evaluations}

{\bf Evaluations on different displacements.}
Our global matching aims at addressing large motions explicitly. To evaluate its performance, we divide all regions of the Sintel training set (both clean and final passes) into different subsets, i.e., \emph{s}10, \emph{s}10-40, \emph{s}40+, based on displacements. \emph{s}10 refers to regions with displacements between 0 and 10, \emph{s}10-40 for 10 and 40, and \emph{s}40+ for larger than 40. Then, we train the optimization-only baseline model RAFT \cite{teed2020raft} and GMFlowNet on C+T and evaluate them on the different subsets. 
Table~\ref{tab:eval_displacement} provides the evaluation results in terms of average end-point-error (AEPE). 
As shown, for the clean pass, GMFlowNet improves RAFT by 22.4\% (from 8.80 from 6.83) on \emph{s}40+ and 18.3\% (from 1.38 from 1.69) on \emph{s}10-40. 
For the final pass, GMFlowNet is close to RAFT on \emph{s}10 and \emph{s}10-40 but outperforms RAFT on \emph{s}40 by 4.7\%. 
Those results indicate that GMFlowNet enjoys great improvements on regions with extremely large displacements, which demonstrates that the global matching with large context information is beneficial to handle large motions.

\begin{table}
  \setlength{\tabcolsep}{0.5\tabcolsep}
  \centering
\begin{tabular}{ll*{3}{c}}
\toprule
Sintel &  & RAFT\cite{teed2020raft} & Ours & \emph{Rel.~Impr.} \\
Dataset & Type & (AEPE) & (AEPE) & (\%)\\
\midrule
& s0-10 & $0.37$ & $\mathbf{0.28}$ &  $24.3$ \\
Clean & \emph{s}10-40 & $1.69$ & $\mathbf{1.38}$ & $18.3$ \\
(train) & \emph{s}40+ & $8.80$ & $\mathbf{6.83}$ & $22.4$ \\
& All & $1.47$ & $\mathbf{1.14}$ & $22.4$ \\
\midrule
& \emph{s}0-10 & ${\bf 0.53}$ & $0.54$ & $-1.9$ \\
Final& \emph{s}10-40 & $3.11$ & $\mathbf{3.09}$ & $0.6$ \\
(train)& \emph{s}40+ & $18.11$ & $\mathbf{17.25}$ & $4.7$ \\
& All & $2.78$ & $\mathbf{2.71}$ & $2.5$\\
\bottomrule
\end{tabular}
\caption{{\bf Quantitative results on different displacements.} Models are trained on C+T. \emph{Rel.~Impr.} refers to relative improvement. Our method improves more on regions with extremely large motions (\emph{s}40+) than on \emph{s}10-40.}
  \label{tab:eval_displacement}
\end{table}

\setlength\tabcolsep{.7em}
\begin{table*}[ht]
\centering
\resizebox{.95\textwidth}{!}{
\begin{tabular}{clccccccc}
\toprule
\multirow{2}{*}{Training} & \multirow{2}{*}{Method} & \multicolumn{2}{c}{\underline{Sintel (train)}} &  \multicolumn{2}{c}{\underline{KITTI-15 (train)}} & \multicolumn{2}{c}{\underline{Sintel (test)}} & \multicolumn{1}{c}{\underline{KITTI-15 (test)}} \\
& & Clean & Final & F1-epe & F1-all & Clean & Final & F1-all \\
\midrule    
\multirow{11}{*}{C+T} 
                       & HD3\cite{yin2019hierarchical}            & 3.84  & 8.77 & 13.17 & 24.0 & - & - & - \\
                       & PWC-Net\cite{sun2018pwc}        & 2.55  & 3.93 & 10.35 & 33.7 & - & - & - \\
                      & LiteFlowNet2\cite{hui2020lightweight}   & 2.24  & 3.78  & 8.97 & 25.9 & - & - & - \\
                       & VCN\cite{yang2019volumetric}            & 2.21  & 3.68  & 8.36 & 25.1 & - & -     & - \\ 
                       & MaskFlowNet\cite{Zhao2020MaskFlownetAF} & 2.25 & 3.61 & - & 23.1 & - & - & - \\ 
                       & FlowNet2\cite{ilg2017flownet}       & 2.02  & \ 3.54 & 10.08 & 30.0 & 3.96  & 6.02 & - \\
                       & DICL-Flow\cite{wang2020displacement} & 1.94 & 3.77 & 8.70 & 23.6 & - & - & - \\
                       & RAFT\cite{teed2020raft} & 1.43 & \underline{2.71} & 5.04 & 17.4 & - & - & -\\
                       & GMA \cite{jiang2021learning} & \underline{1.30} & 2.74 & 4.69 & 17.1 & - & - & -\\
                       & Separable Flow\cite{zhang2021separable} & \underline{1.30} & {\bf 2.59} & \underline{4.60} & \underline{15.9} & - & - & -\\
                        & {\bf GMFlowNet (Ours)} & {\bf 1.14}  & \underline{2.71} & {\bf 4.24} & {\bf 15.4} & - & - & - \\
                       \midrule
\multirow{12}{*}{C+T+S/K} & FlowNet2 \cite{ilg2017flownet}  & (1.45) & (2.01) & (2.30) & (6.8) & 4.16  & 5.74 & 11.48  \\
                     & HD3 \cite{yin2019hierarchical}         & (1.87)     & (1.17) & (1.31) & (4.1)  & 4.79  & 4.67 & 6.55 \\
                     & PWC-Net\cite{sun2018pwc}        & -  & - & - & - & 4.39 & 5.04 & 9.60 \\
                     & LiteFlowNet\cite{hui2018liteflownet} & (1.35) & (1.78) & (1.62) & (5.58) & 4.54 & 5.38 & 9.38 \\
                     & ScopeFlow\cite{bar2020scopeflow} & - & - & - & - & 3.59 & 4.10 & 6.82 \\
                     & VCN \cite{yang2019volumetric}            & (1.66)     & (2.24) & (1.16) & (4.1) & 2.81  & 4.40 & 6.30 \\
                     & DICL-Flow\cite{wang2020displacement} & (1.11) & (1.60) & (1.02) & (3.60) & 2.12 & 3.44 & 6.31 \\
                     & RAFT*\cite{teed2020raft} & (0.77) & (1.20) & (0.64) & (1.5) & 2.08 & 3.41 & 5.27\\
                     & Separable Flow\cite{zhang2021separable} & (0.71) & (1.14) & (0.68) & (1.57) & \underline{1.99} & \underline{3.27} & {\bf 4.89}\\
                     & {\bf GMFlowNet (Ours)} & (0.65) & (1.06) & (0.63) & (1.49) & {\bf 1.59} & {\bf 2.91} & {\bf 4.89} \\ 
                     \midrule
                     
\multirow{8}{*}{C+T+S+K+H}
                     & LiteFlowNet2 \cite{hui2020lightweight} & (1.30) & (1.62) & (1.47) & (4.8) & 3.48  & 4.69 & 7.74 \\
                     & PWC-Net+\cite{sun2019models}   & (1.71)     & (2.34)  & (1.50) & (5.3)  & 3.45  & 4.60 & 7.72 \\
                     & MaskFlowNet\cite{Zhao2020MaskFlownetAF} & - & - & - & - & 2.52 & 4.17 & 6.10 \\
                     & RAFT*\cite{teed2020raft}      & (0.76)  & (1.22) & (0.63) & (1.5) & 1.94 & 3.18 & 5.10 \\ 
                     & GMA* \cite{jiang2021learning} & (0.62) & (1.06) & (0.57) & (1.2) & \underline{1.40} & 2.88  & 5.15 \\
                     & Separable Flow\cite{zhang2021separable} & (0.69) & (1.10) & (0.69) & (1.60) & 1.50 & \underline{2.67}  & {\bf 4.64} \\
                     & {\bf GMFlowNet (Ours)} & (0.59) & (0.91) & (0.64) & (1.51) & {\bf 1.39} & {\bf 2.65} &
                     \underline{4.79} \\ 
                     \bottomrule
\end{tabular}
}
\caption{{\bf Quantitative results on Sintel and KITTI datasets.} 
``C+T'': We test the generalization ability on Sintel and KITTI training sets after training on FlyingChairs (C) and FlyingThing (T). 
``C+T+S/K'': We train models on C+T and finetune them on either Sintel (S) or KITTI (T) and evaluate on the test set of S or T. 
``C+T+S+K+H'': Our training set contains training samples from C, T, S, K and HD1K (H). 
Parentheses denote results on the training set. The best and runner up results are highlighted in bold and underlined, respectively.
*We report results of the 2-view setting that is adopted by other methods.}
\label{table:sota_com}
\end{table*}

{\bf Cross-domain evaluations.}
Following previous studies \cite{teed2020raft,jiang2021learning,zhang2021separable}, we trained the proposed GMFlowNet on C+T and evaluated it on the training sets of Sintel and KITTI as cross-domain evaluations. Table~\ref{table:sota_com} displays the results of GMFlowNet and other competitive approaches. As a common practice, AEPE is reported for Sintel. Fl-epe and Fl-all are reported for KITTI.

As shown, GMFlowNet is close to the best method Separable Flow \cite{zhang2021separable} on Sintel Final and achieves better performance on the other datasets. 
Our method achieves an AEPE of 1.14 on Sintel Clean, a Fl-all of 15.4 on KITTI, which are 19.6\% and 11.5\% better than the optimization-only baseline, RAFT. 
Those results demonstrate that GMFlowNet boasts a better generalization ability than RAFT as well as other methods. 
Considering that GMFlowNet and RAFT share the same optimization stage, we attribute the huge improvement in generalization to our global matching. 
We believe this is a fair claim because RAFT exploits regression but GMFlowNet considers both matching and regression. Since regression is more likely to overfit specific datasets than matching, GMFlowNet generalizes better.

{\bf Evaluations on standard benchmarks.}
We evaluate GMFlowNet on standard online benchmarks, i.e., Sintel \cite{butler2012naturalistic} and KITTI \cite{menze2015object}. For a fair comparison, we follow previous methods \cite{teed2020raft,jiang2021learning,zhang2021separable} and train GMFlowNet on C+T+S/K and C+T+S+K+H, respectively.
Table~\ref{table:sota_com} exhibits the evaluation results.
GMFlowNet adopts the optimization process of RAFT, but outperforms RAFT by a large margin. 
Moreover, GMFlowNet outperforms the state-of-the-art method Separable Flow \cite{zhang2021separable} on Sintel, but achieves slightly lower performance on KITTI. 
This is probably because GMFlowNet adopts attention blocks to extract large context features. However, KITTI only provides 200 training images that are far from enough to train high quality attention blocks. 
We assume that with more training data, GMFlowNet may result in larger improvements compared to CNN-based approaches.


\begin{figure*}
  \centering
  \begin{subfigure}{0.33\linewidth}
      \includegraphics[width=1\linewidth]{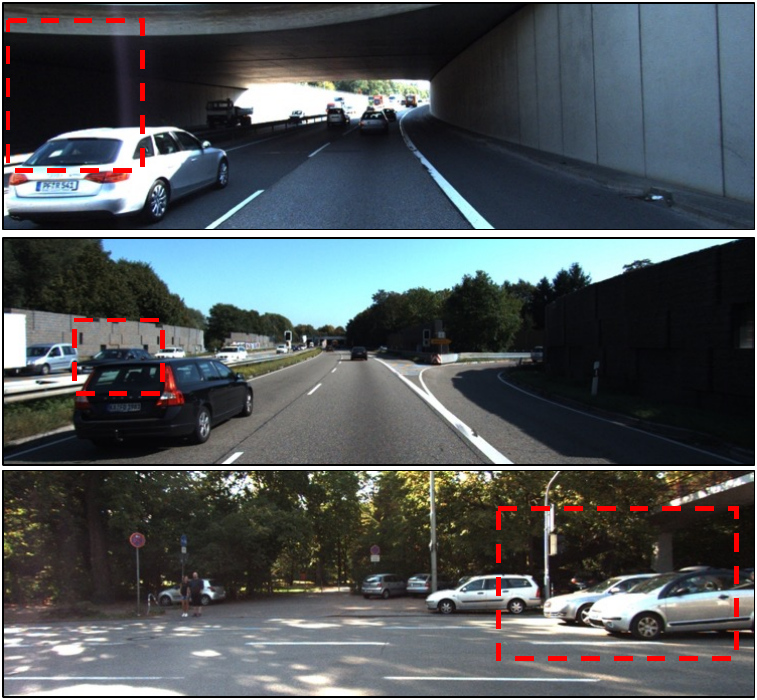}
      \caption{Input images}
      \label{fig:kitti_input}
  \end{subfigure}
  \begin{subfigure}{0.33\linewidth}
      \includegraphics[width=1\linewidth]{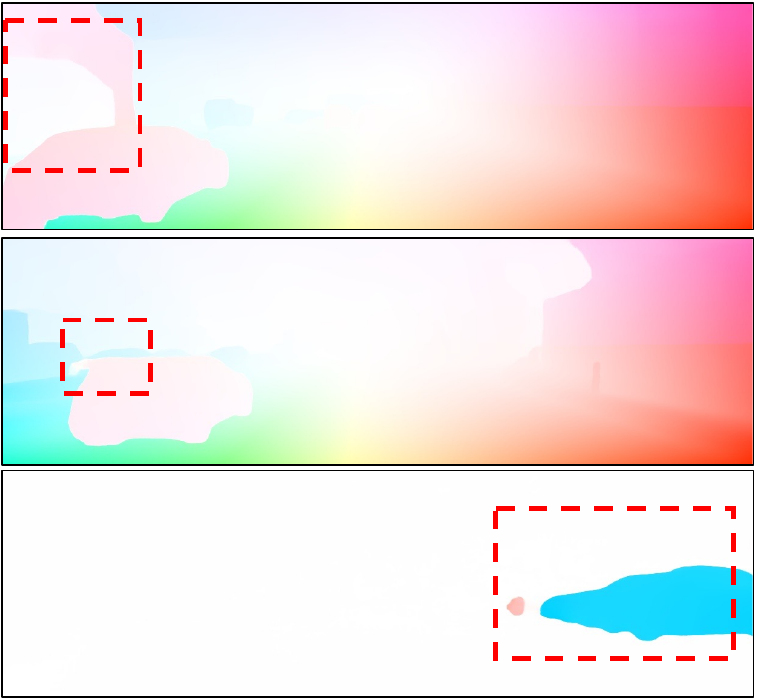}
      \caption{RAFT\cite{teed2020raft}}
      \label{fig:kitti_raft}
  \end{subfigure}
  \begin{subfigure}{0.33\linewidth}
      \includegraphics[width=1\linewidth]{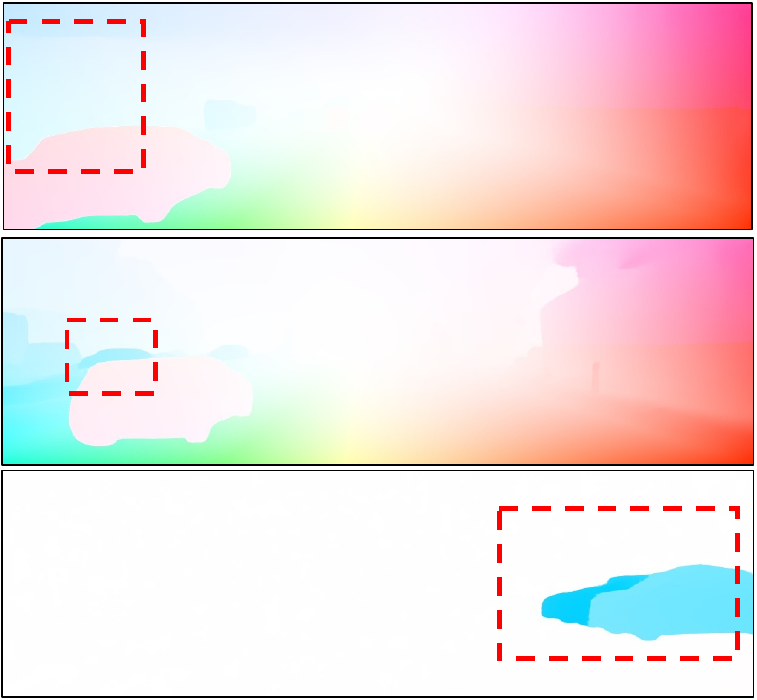}
      \caption{Ours}
      \label{fig:kitti_outs}
  \end{subfigure}
  \caption{{\bf Qualitative evaluations} for four samples from KITTI test set. (b) Results of the widely adopted optimization-only baseline model RAFT \cite{teed2020raft}. (c) Results of our GMFlowNet. Regions with significant improvements are highlighted by red dash boxes. GMFlowNet works better especially in textureless regions, because our overlapping attention provides more context information to diminish regional ambiguities.} \label{fig:kitti_vis}
\end{figure*}

\subsection{Qualitative Evaluations}

We visualize the estimated flows and cost volumes to illustrate the exact aspects that GMFlowNet improves. The supplementary document provides additional visualizations for ours coarse flow from matching.

{\bf Visualizations of estimated flows.} Fig.~\ref{fig:kitti_vis} provides several test samples from KITTI and the corresponding flow estimations of RAFT and GMFlowNet. As we can see, compared with RAFT, GMFlowNet provides better predictions on locally ambiguous regions like textureless regions. For example, there are two white cars moving forward side by side in the last row of Fig.~\ref{fig:kitti_vis}. Since the two cars share similar colors and shapes, RAFT interprets them as one car. In contrast, our method succeeds in estimating the difference between the two cars and predicts the flow correctly. 
For more results, please refer to The supplementary Sect~\ref{sect:more_visual}.
These improvements are strong evidence of the effectiveness of the introduced global matching and POLA.

\begin{figure}[t]
  \centering
  \begin{subfigure}{0.49\linewidth}
    \centering
        \includegraphics[width=1\linewidth, trim=0 40 0 40]{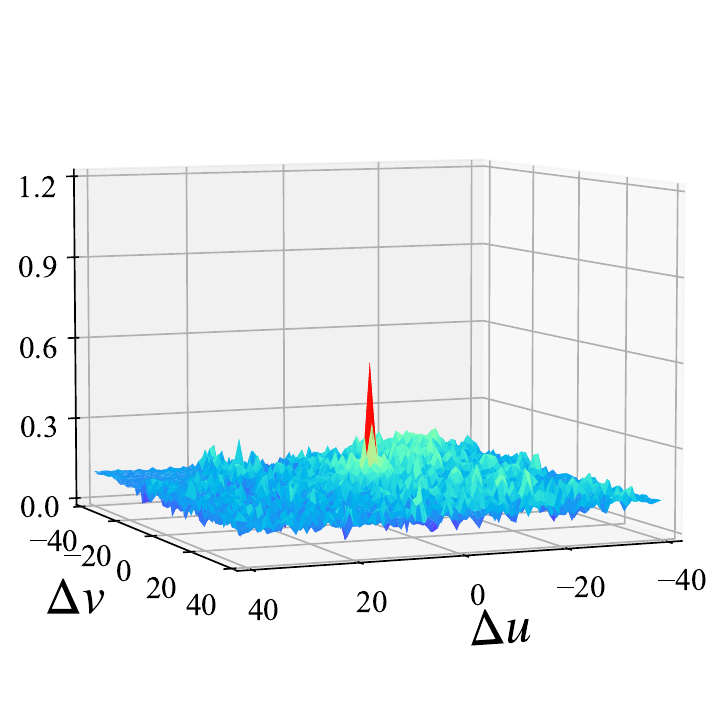}
    \caption{RAFT\cite{teed2020raft}}
    \label{fig:cost_comp_raft}
  \end{subfigure}
  \begin{subfigure}{0.49\linewidth}
    \centering
        \includegraphics[width=1\linewidth, trim=0 40 0 40]{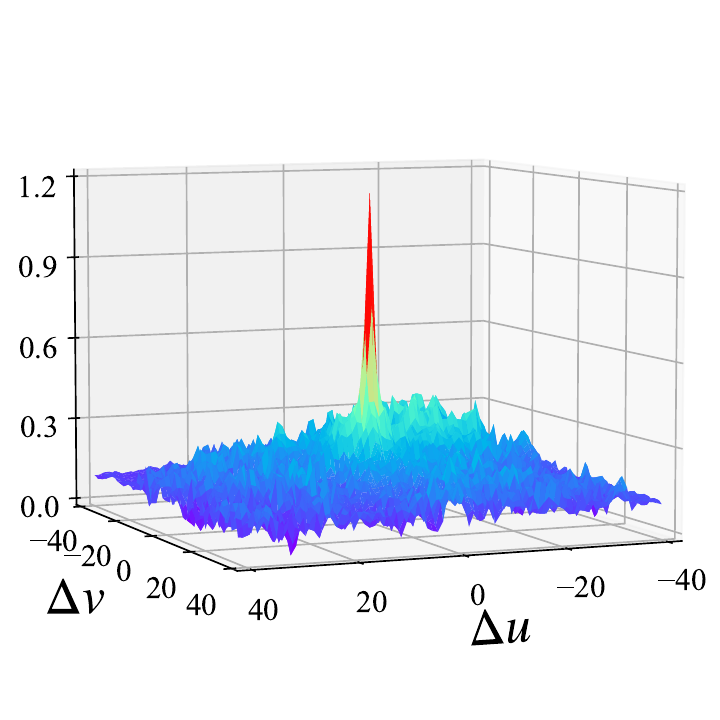}
    \caption{Ours}
    \label{fig:cost_comp_ours}
  \end{subfigure}
  \caption{\textbf{Visualizations of cost volumes for large motions.} 
  The peak of our cost volume is twice higher than that of RAFT's, which demonstrates that our method handles large displacements better.
  }
  \label{fig:cost_comp}
\end{figure}

{\bf Visualizations of cost volumes.} Fig.~\ref{fig:cost_comp} visualizes the average and normalized cost volumes of both RAFT and GMFlowNet for large displacement regions ($>$ 20 pixels). The supplementary Sect~\ref{sect:more_visual} provides more details about the visualization. 
For a fair comparison, we trained RAFT and GMFlowNet on C+T and drew the figure using the training set of Sintel. 
As shown, the peak of our cost volume is much higher than that of RAFT, which clearly demonstrates that GMFlowNet is better at handling large displacements. 
This is plausible because our matching is designed to handle large motions and the proposed POLA extracts large context information that is crucial to overcome regional ambiguities for matching.

\setlength\tabcolsep{.7em}
\begin{table*}[t]
\centering
\resizebox{0.95\textwidth}{!}{
\begin{tabular}{clccccc}
\toprule
\multirow{2}{*}{Experiment} & \multirow{2}{*}{Method} & \multicolumn{2}{c}{\underline{Sintel (train)}} & \multicolumn{2}{c}{\underline{KITTI-15 (train)}} & \multirow{2}{*}{Parameters} \\
& & Clean & Final & F1-epe & F1-all &  \\
\midrule 
Baseline \cite{teed2020raft} & - & 1.43 & 2.71 & 5.04 & 17.4 & 5.3M  \\
\midrule
\multirow{3}{*}{Initial Features}  
                            & None & 1.31 & 2.83 & 4.89 & 17.4 & 11.1M \\
                            & ResNet & 1.27 & 2.86 & 4.44 & 16.3 & 10.0M \\
                            & \underline{3-Convs} & 1.14 & 2.71 & 4.24 & 15.4 & 9.3M \\
                            \midrule
\multirow{4}{*}{Large Context Features}      
                            & ResNet & 1.26 & 2.95 & 4.74 & 17.1 & 5.3M \\
                            & Swin Transformer & 1.33 & 2.90 & 5.65 & 17.3 & 9.3M \\ 
                            & ViT & 1.33 & 2.94 & 4.86 & 16.4 & 12.5M \\ 
                            & \underline{POLA} & 1.14 & 2.71 & 4.24 & 15.4 & 9.3M \\
                            \midrule
\multirow{3}{*}{Number of Attention Blocks}    
                            & 3 & 1.27 & 2.87 & 4.76 & 16.9 & 6.9M \\
                            & \underline{6} & 1.14 & 2.71 & 4.24 & 15.4 & 9.3M \\
                            & 12 & 1.12 & 2.63 & 4.04 & 15.6 & 14.1M \\ 
                            \midrule
\multirow{2}{*}{Overlapping Type}    
                            & Per pixel & 1.32 & 2.88 & 5.11 & 16.8 & 9.3M \\
                            & \underline{Patch-based} & 1.14 & 2.71 & 4.24 & 15.4 & 9.3M \\ 
                            \midrule
\multirow{2}{*}{Global Matching}        
                            & No & 1.24 & 2.82 & 4.58 & 16.4 & 9.3M \\
                            & \underline{Yes} & 1.14 & 2.71 & 4.24 & 15.4 & 9.3M \\ 
\bottomrule
\end{tabular}
}
\caption{{\bf Ablation experiments.} Settings used in the final model are underlined. See Sec.~\ref{sect:ab_study} for details.}
\label{table:Ablations}
\end{table*}

\subsection{Ablation Study}\label{sect:ab_study}

We perform a set of ablation studies to show the importance and effectiveness of each component in GMFlowNet. All models in the experiments are trained on C+T and tested on Sintel and KITTI training sets. Table~\ref{table:Ablations} provides the results for various ablation experiments. In each section of the table, we study a specific component of our approach in isolation and underline the settings used in our final model.

{\bf Initial feature extraction.} We tried three different modules, i.e., None, ResNet \cite{he2016deep}, and 3-Convs, to extract initial features for the following POLA blocks. None means no initial features. For this setting, we use the Swin Transformer architecture as the overall feature extractor,  but we replace its attention blocks with POLA blocks.
As shown in Table~\ref{table:Ablations}, 3-Convs achieve the best performance. This is likely because, on the one hand, attention blocks have more difficulties than CNNs to learn rich features from raw images. On the other hand, ResNet is much deeper than 3-Convs and may extract more high level features that are less useful for matching.

{\bf Large context feature extraction.} To verify the effectiveness of our POLA, we compare it with ResNet, Swin Transformer \cite{liu2021swin}, and ViT \cite{dosovitskiy2021an}. For ViT, we further reduce the feature maps by 4x. Otherwise, ViT will run out of memory because it takes global attentions instead of local attentions used in POLA.
As shown in Table~\ref{table:Ablations}, POLA outperforms others by a large margin.

{\bf The number of our attention blocks.} A simple way to expand GMFlowNet is to increase the number of attention blocks. Table~\ref{table:Ablations} shows that more attention blocks achieve better performance, which is probably because more blocks provide larger receptive fields and better context information. However, more blocks increases the computation and memory costs. As a trade-off, we take 6 blocks finally.

{\bf Overlapping type.} Our POLA can be viewed as a generalization of per pixel overlapping attention proposed in \cite{ramachandran2019stand}. As shown in Table~\ref{table:Ablations}, POLA shares the same amount of parameters with the per pixel attention and outperforms it.

{\bf Global matching.} Our key motivation is to introduce global matching into direct-regression methods. We remove the global matching in GMFlowNet and observe a significant performance drop on Sintel Final and KITTI shown in Table~\ref{table:Ablations}. 
Those results clearly demonstrate the effectiveness of the global matching.

\begin{table}
  \centering
  \begin{tabular}{l*{4}{c}}
    \toprule
    \multirow{2}{*}{{\bf Method}} & \multirow{2}{*}{{\bf Param}} & \multirow{2}{*}{{\bf Speed}} & {\bf Sintel} & {\bf KITTI}
    \\
    & & & {\bf Clean} & {\bf Fl-epe}  \\
    \midrule
    RAFT\cite{teed2020raft} & 5.3M & 0.382s & 1.43 & 5.04\\
    RAFT+GM & 5.3M & 0.384s & 1.26 & 4.74 \\
    \midrule
    +SWIN\cite{liu2021swin} & 9.3M & 0.422s & 1.33 & 5.65 \\
    Ours & 9.3M & 0.500s & 1.14 & 4.24\\
    \bottomrule
  \end{tabular}
  \caption{{\bf Comparisons of parameters and inference time.} All models are trained on C+T and tested on S and K. Speed measurements are evaluated on Sintel with the same platform.}
  \label{tab:param_time}
\end{table}

\subsection{Efficiency}

{\bf Running time cost of our global matching.} Running time cost is a major concern to adopt global matching. To address this concern, we compare the running time of the widely adopted RAFT and RAFT+GM. RAFT+GM refers to RAFT with our global matching step. As shown in Table~\ref{tab:param_time}, the global matching is very efficient and only takes 0.002s or 0.52\% of extra time. 
Moreover, compared with RAFT, GMFlowNet runs slightly slower with 4M more parameters but significantly improves the performance. Therefore, the main benefit of our method is the performance improvement.

{\bf Running time cost of our overlapping attention.} Our overlapping attention introduces more calculations but is not necessarily inefficient. To demonstrate this, we compare our model with +SWIN in Table~\ref{tab:param_time}. +SWIN is a variant model where the POLA blocks are replaced with local attention blocks from Swin Transformer \cite{liu2021swin}. As shown, compared with +Swin, GMFlowNet requires 0.078s of extra time and improves the performance by 13.5\% on Sintel clean pass and by 24.9\% on KIITI. 
We believe that the overhead is acceptable given the performance improvement.

\section{Conclusion}
We have shown that matching improves the performance of  direct-regression optical flow estimation methods in handling large displacements. We proposed a novel framework, GMFlowNet, where a global matching step is introduced before learning-based optimization. To improve the matching, we proposed a patch-based overlapping attention that extracts large context features to diminish regional ambiguities.
GMFlowNet significantly improves predictions for large motions and textureless regions and achieves state-of-art performance on standard benchmark datasets. Future work may focus on addressing GMFlowNet's limitations on running time cost and number of parameters.

\vspace{1mm}
\noindent {\bf Acknowledgments.} 
We thank Samuel Schulter from NEC Laboratories America for helpful discussions. 
This research has been partially funded by the following grants, NSF IUCRC CARTA, ARO MURI 805491, NSF IIS-1793883, NSF CNS-1747778, NSF IIS 1763523, DOD-ARO ACC-W911NF, NSF OIA-2040638 to Dimitris Metaxas.

{\small
\bibliographystyle{ieee_fullname}
\bibliography{egbib}
}

\clearpage


\section{Architecture Details}
\label{sec:arch}

\subsection{Large Context Feature Extraction}

In the paper, we exploit Transformer blocks to extract large context features to improve the matching step in GMFlowNet.
In the original Transformer block \cite{vaswani2017attention}, input features are updated by a Multi-head Self-Attention (MSA) followed by a Multilayer perceptron (MLP). 
MSA is able to extract the long-term dependency, and MLP projects the features to the required dimension. 
Both MSA and MLP calculate residuals that are added to the input features as the output features. The update in a transformer block can be formulated as,
\begin{align}
    \hat{x}^{l} &= \text{MSA}(\text{LN}(x^{l-1})) + \hat{x}^{l} \nonumber\\
    x^{l} &= \text{MLP}(\text{LN}(\hat{x}^{l})) + \hat{x}^{l},
\end{align}
where LN refers to layer norm, and $x^{l-1}$ and $x^{l}$ represent output features of the previous block and the current block, respectively.
The MSA is originally designed for language tasks and takes the whole 1D features as input, but it is computationally prohibitive to apply it on 2D feature maps for optical flow estimation. 
To extract the long-term dependency with an acceptable computation cost, we propose the patch-based overlapping attention (POLA) to replace MSA of the original attention block and call our attention block as multi-head POLA (M-POLA).

In our large context feature extraction module (Section 3.1), we take 3 convolutional layers (3-Convs) to extract initial features and 6 M-POLA blocks to extract large context information based on initial features. The detailed structure of this module is listed in Table~\ref{tab:ctx_feats}.

\subsection{Optimization Network}

We adopt the iterative update operator proposed in RAFT \cite{teed2020raft} as the optimization step of GMFlowNet. As stated in \cite{teed2020raft}, this operator mimics the steps of an optimization algorithm and iteratively outputs a series of flow predictions $\{f_{1\to 2}^{(1)},f_{1\to 2}^{(2)},\dots,f_{1\to 2}^{(T)}\}$.
For the $t$-th iteration, the flow prediction $f_{1\to 2}^{(t)}$ is calculated by a Convolutional GRU \cite{cho2014properties} (ConvGRU) as,
\begin{equation}
\begin{aligned}
x^{(t)} &= [f^{(t-1)}_{1\to2}, F_1, \text{lookup}(C, f^{(t-1)}_{1\to2}, r)],\\
r^{(t)} &= \sigma(\text{Conv}([h^{(t-1)}, x^{(t)}])),\\
{\widetilde h}^{(t)} &=   \sigma(\text{Conv}([r^{(t)} \odot h^{(t-1)}, x^{(t)}])),\\
z^{(t)} &= \mu(\text{Conv}([h^{(t-1)}, x^{(t)}])),\\
h^{(t)} &= (1 - z^{(t)}) \odot h^{(t - 1)} + z^{(t)} \odot {\widetilde h}^{(t)},\\
\Delta f^{(t)}_{1\to2} &= \text{Conv}(h^{(t)}), \\
f^{(t)}_{1\to2} &= f^{(t-1)}_{1\to2} + \Delta f^{(t)}_{0\to1}
\end{aligned}
\label{eq:conv_gru}
\end{equation}
where $F_1$ is the context features, $C$ is the 4D cost volume (See Section 3.2 of the paper), $\text{Conv}(\cdot)$ refers to a convolution layer, $\sigma(\cdot)$ means sigmoid, and $\mu(\cdot)$ means tanh. $\text{lookup}(\cdot)$ represents the cost volume within the range of $r$. For each location $\mathbf{x}$ in $I_1$, $\text{lookup}(\cdot)$ is defined as,
\begin{align}
\text{lookup}(\cdot) = \{C(\mathbf{x}, f^{(t-1)}_{1\to2}(\mathbf{x}) + \mathbf{\delta x} )
\mid 
r > \parallel \mathbf{\delta x} \parallel_1 \} .
\end{align}
Different iterations share the weights in the ConvGRU.

\begin{table}[t]
    \centering
    \begin{tabular}{c|l}
     \bottomrule
     & \bf{Layer name(s)} \\
    \hline \hline
    \multirow{3}*{\rotatebox{90}{3-Convs}} 
    & Conv(3, 64, 7, 2), ReLU \\
    & Conv(64, 128, 3, 2), ReLU \\
    & Conv(128, 256, 3, 2), ReLU \\
    \hline
    \multirow{6}*{\rotatebox{90}{6 M-POLA}} 
    & M-POLA (dim=256, head=8, win\_size=7) \\
     & M-POLA (dim=256, head=8, win\_size=7) \\
    & M-POLA (dim=256, head=8, win\_size=7) \\
    & M-POLA (dim=256, head=8, win\_size=7) \\
     & M-POLA (dim=256, head=8, win\_size=7) \\
    & M-POLA (dim=256, head=8, win\_size=7) \\
    \toprule
    \end{tabular}
    \caption{Large context feature extraction. The arguments in Conv($\cdot$) are the input channel number, the output channel number, the kernel size, and the convolution stride, respectively.} \label{tab:ctx_feats}
\end{table}

\begin{figure*}[t]
  \centering
  \includegraphics[width=1.0\linewidth]{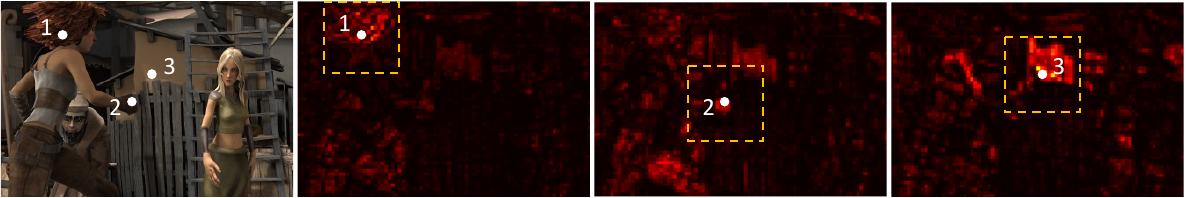}
   \caption{Visualization of attention scores. The more red a pixel is, the higher the score is.}\label{fig:attn_vis}
\end{figure*}

\section{More Visualizations}\label{sect:more_visual}

\subsection{Attention maps}
Fig.~\ref{fig:attn_vis} visualizes full attention score maps of the first POLA for three pixels highlighted in white. The more red a pixel is, the higher the score is. Yellow dash boxes indicate the local regions that are used in POLA. As shown, a pixel is more likely to attend to those that are visually similar to the pixel.

\subsection{Coarse Flows}

\begin{figure*}
  \centering
  \begin{subfigure}{1\linewidth}
      \includegraphics[width=1\linewidth]{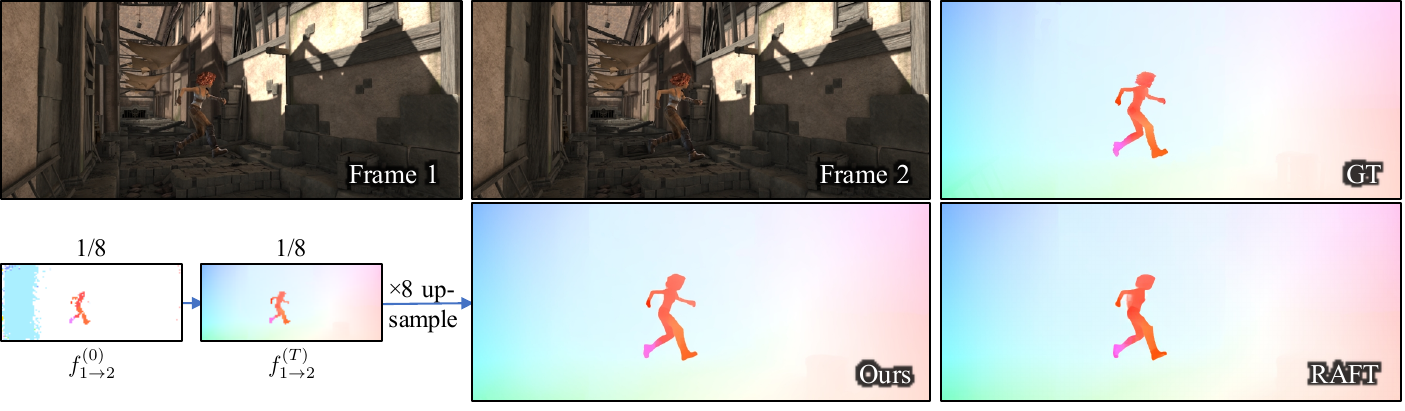}
      \includegraphics[width=1\linewidth]{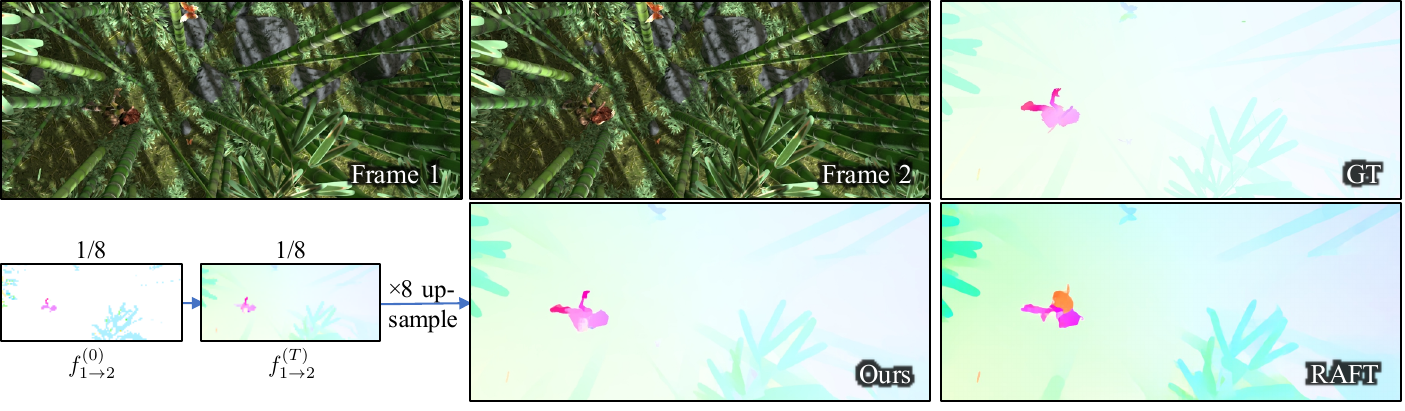}
      \caption{Samples from Sintel training set \cite{butler2012naturalistic}}
      \label{fig:sintel_gmflow}
  \end{subfigure}
  \begin{subfigure}{1\linewidth}
      \includegraphics[width=1\linewidth]{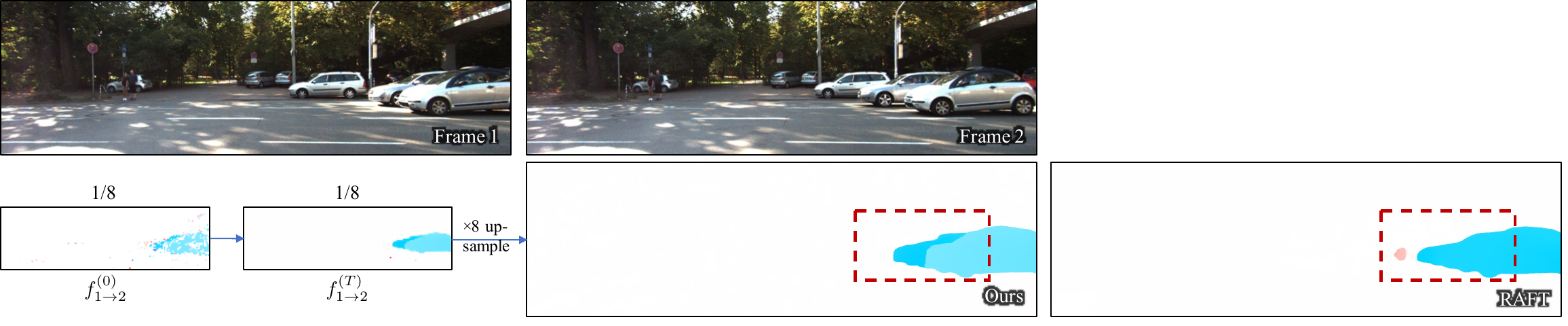}
      \includegraphics[width=1\linewidth]{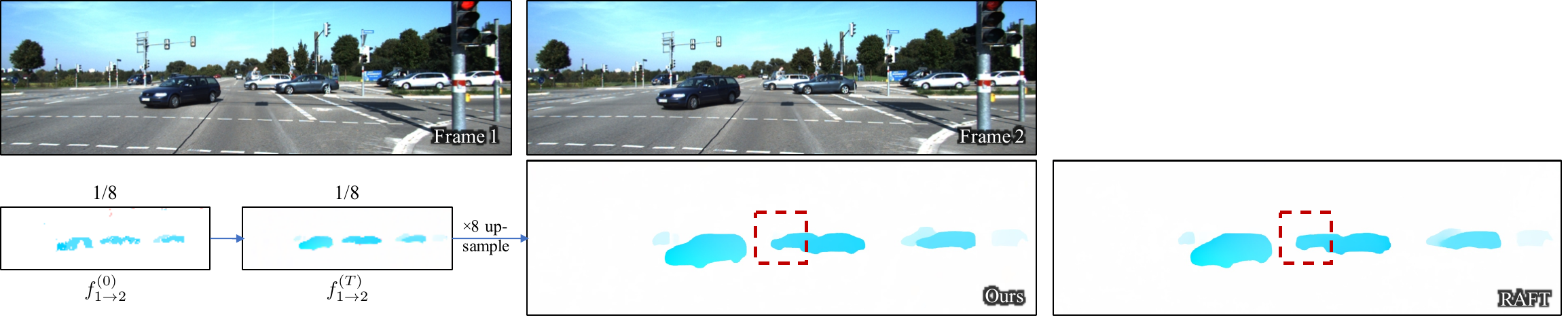}
      \caption{Samples from KITTI test set \cite{menze2015object}}
      \label{fig:kitti_gmflow}
  \end{subfigure}
  \caption{{\bf Visualizations of coarse flow.} For (a) Sintel, models are trained on C+T. For (b) KITTI, models are trained on C+T+S+K+H. Ground-truth flows for KITTI are unavailable and thus are not shown. 
  With the coarse flow, our method outperforms the most popular optimization-only method RAFT \cite{teed2020raft}. Red dash boxes highlight the main differences between RAFT's predictions and ours.
  }  \label{fig:gmflow}
\end{figure*}

Figure~\ref{fig:gmflow} displays the coarse flows from our matching step as well as the final flow estimation for samples from Sintel \cite{butler2012naturalistic} and KITTI \cite{menze2015object} datasets. 
We compare our GMFlowNet with RAFT \cite{teed2020raft} because they share the same optimization architecture. 
For Sintel, both models are trained on C+T. 
For KITTI, they are trained on all the training data. 
As shown, the coarse flow results in better predictions especially in large motion areas and textureless regions. For example, the hand of the character in Fig.~\ref{fig:kitti_gmflow} moves fast, leading to failures of RAFT. On the contrary, our matching step finds the optical flow for the hand and improves the final prediction.

\subsection{More Visual Results}

\begin{figure*}
  \centering
  \begin{subfigure}{1\linewidth}
      \includegraphics[width=.33\linewidth]{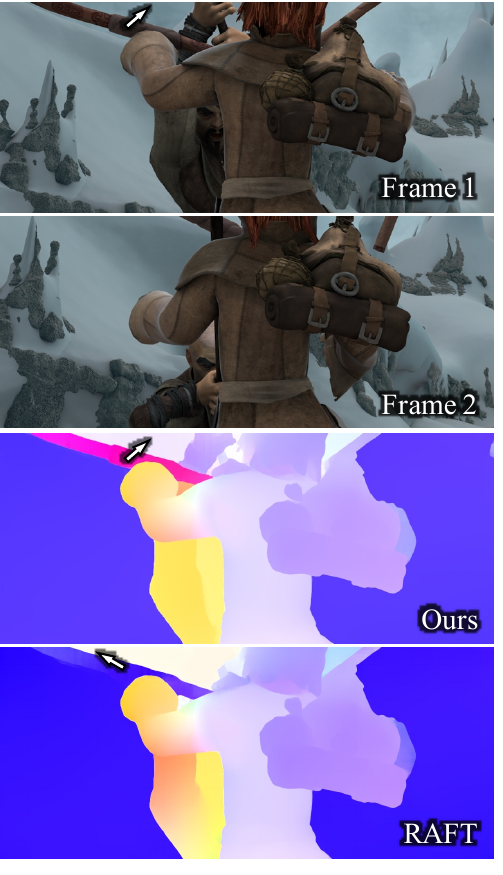}
      \includegraphics[width=.33\linewidth]{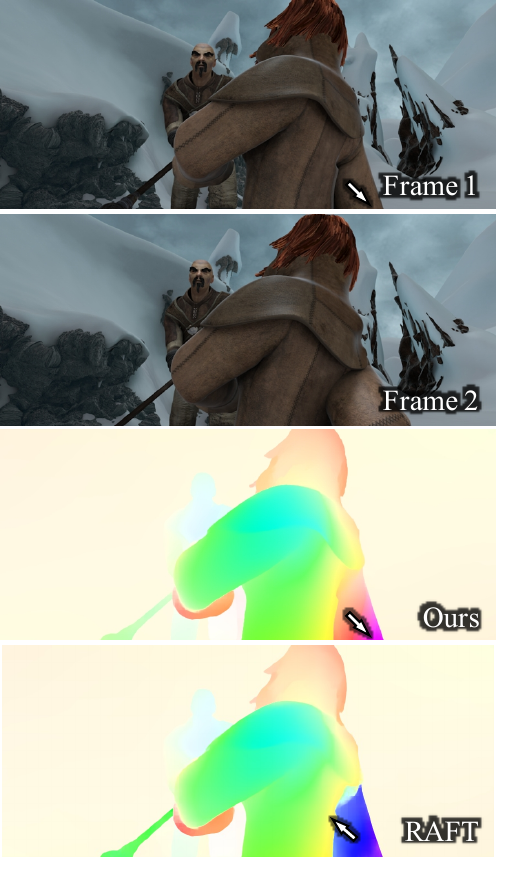}
      \includegraphics[width=.33\linewidth]{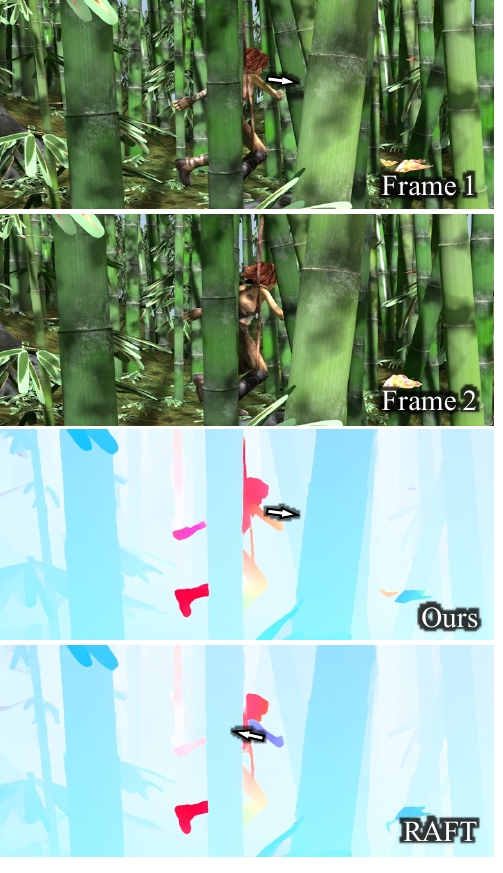}
      \caption{}
  \end{subfigure}
  \begin{subfigure}{1\linewidth}
      \includegraphics[width=.33\linewidth]{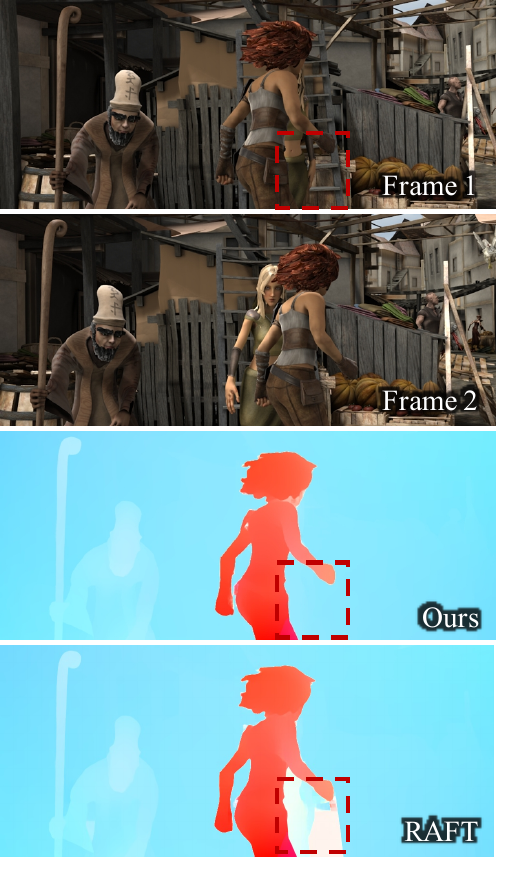}
      \includegraphics[width=.33\linewidth]{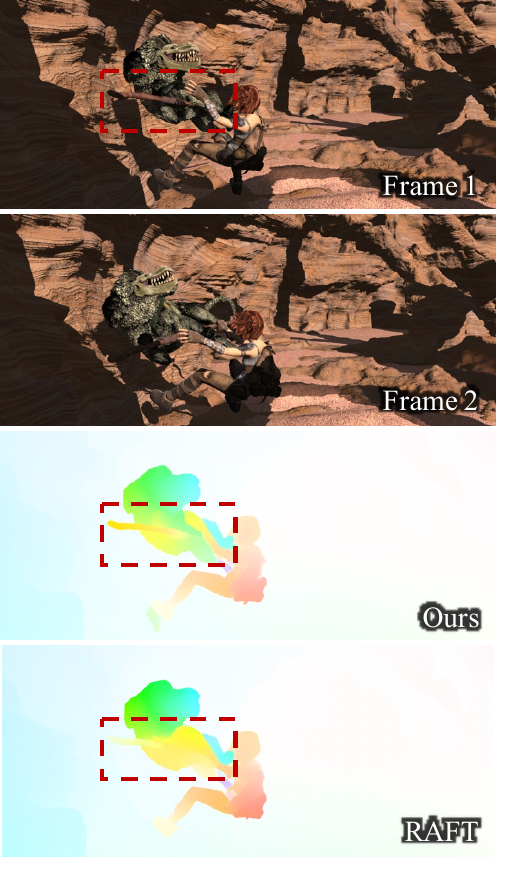}
      \includegraphics[width=.33\linewidth]{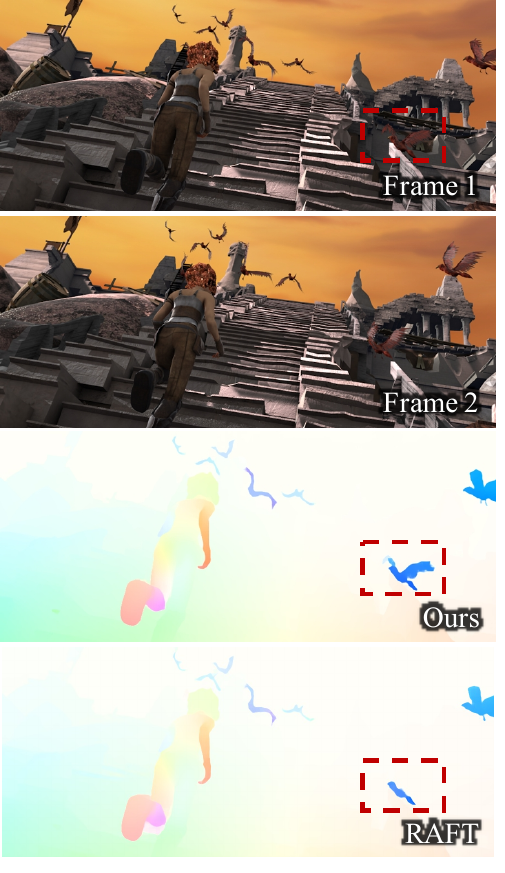}
      \caption{}
  \end{subfigure}
  \caption{{\bf Qualitative evaluation} on the Sintel test set \cite{butler2012naturalistic}. White arrows in (a) and red dash boxes in (b) highlight the differences between our method and RAFT. Ground-truth optical flows are not available and are not shown. Models are trained on the same training data.}  \label{fig:more_sintel}
\end{figure*}

\begin{figure*}
  \centering
  \begin{subfigure}{1\linewidth}
      \includegraphics[width=.33\linewidth]{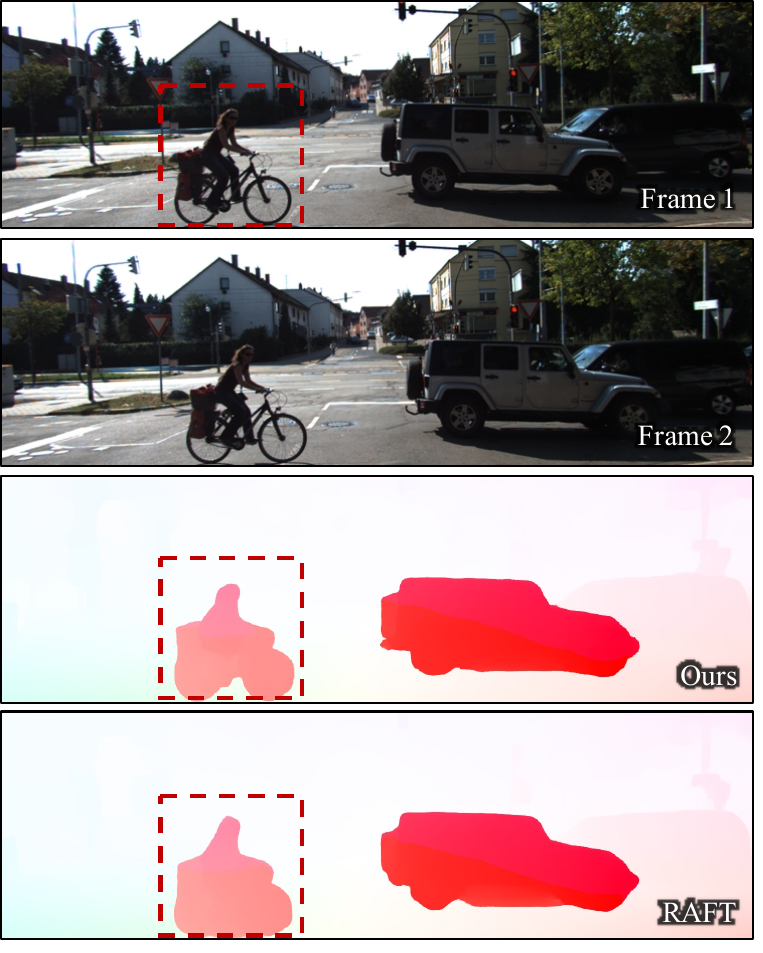}
      \includegraphics[width=.33\linewidth]{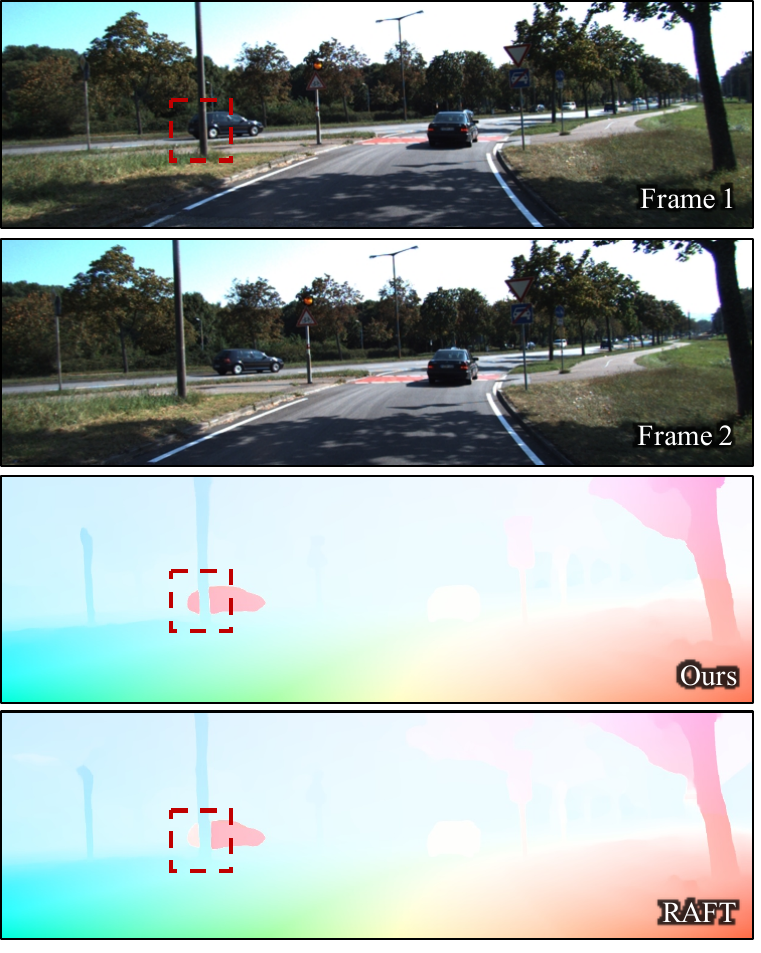}
      \includegraphics[width=.33\linewidth]{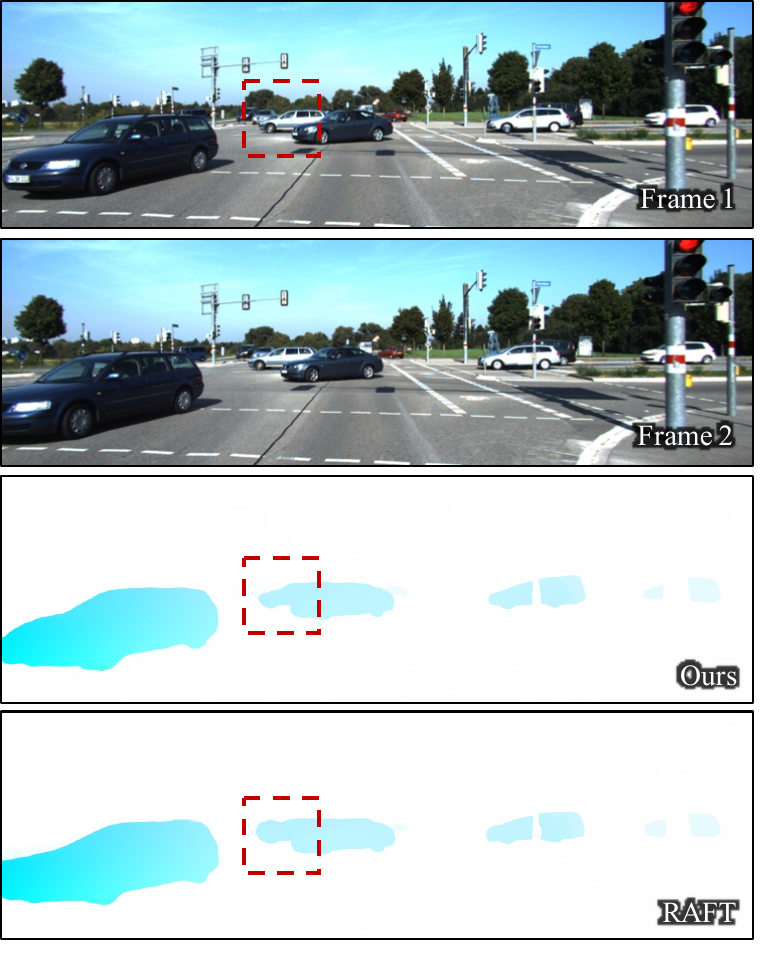}
      \caption{}
  \end{subfigure}
  \caption{{\bf Qualitative evaluation} on the KITTI test set \cite{menze2015object}. Red dash boxes highlight the differences between our method and RAFT. Models are trained on the same training data.}  \label{fig:more_kitti}
\end{figure*}

Figure~\ref{fig:more_sintel} provides the qualitative evaluation of GMFlowNet and RAFT on the Sintel test set. We highlight with white arrows and red dash boxes the regions where our method outperforms RAFT.
Fig.~\ref{fig:more_kitti} exhibits the visualization of more samples from the KITTI test set. Red dash boxes highlights the regions where our method outperforms RAFT.

\section{How We Visualize Cost Volumes}

In order to compare the $4D$ cost volumes $C$ of RAFT~\cite{teed2020raft} and our method, we extract the matrix $F_{x, y}$ as the matching matrix for the point $(x, y)$, 
\begin{align}
    F_{x, y} = \text{softmax}(C[x, y, &(x + \delta x - 40): (x + \delta x + 40), \nonumber \\ 
    &(y + \delta y - 40):( y + \delta y + 40)])
\end{align}
where $\delta x$ and $\delta y$ are indicated by the ground truth flow at $(x, y)$. The symbol $C[\cdot]$ means to fetch values from $C$ within a given range.
Then, we average $F_{x, y}$ on all points within a specific displacement range for all images in Sintel and visualize the averaged matching matrix. 

\begin{figure*}
  \centering
  \begin{subfigure}{0.24\linewidth}
    \includegraphics[width=1\linewidth, trim=0 40 0 40]{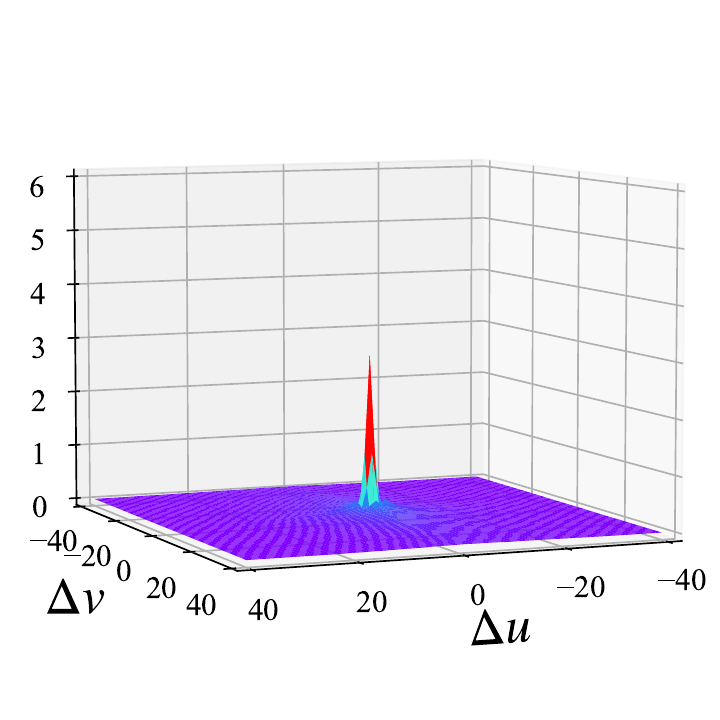}
    \includegraphics[width=1\linewidth, trim=0 40 0 0]{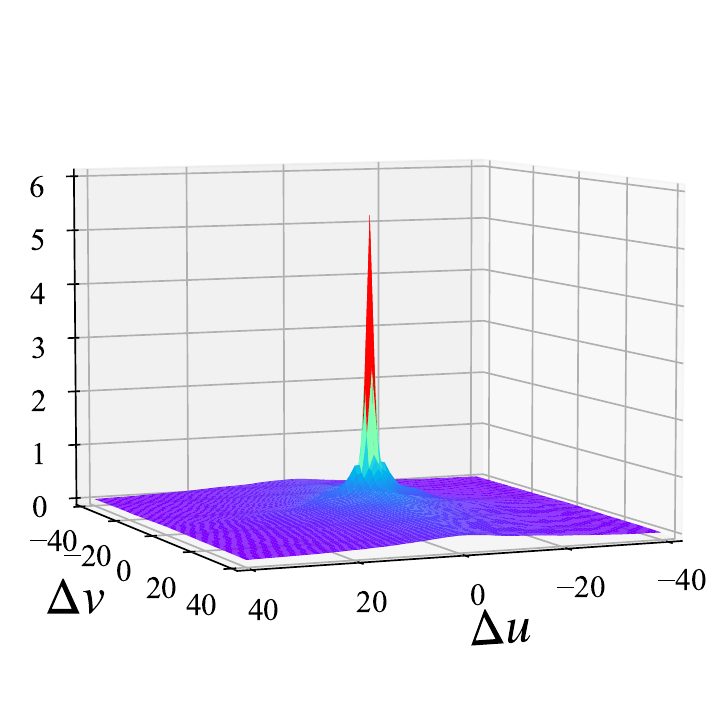}
    \caption{$s10$}
  \end{subfigure}
  \begin{subfigure}{0.24\linewidth}
    \includegraphics[width=1\linewidth, trim=0 40 0 40]{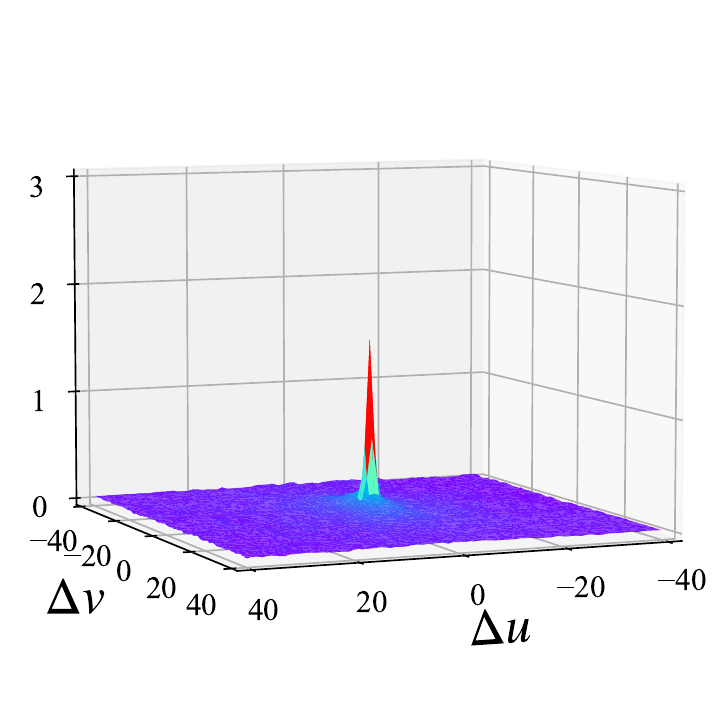}
    \includegraphics[width=1\linewidth, trim=0 40 0 0]{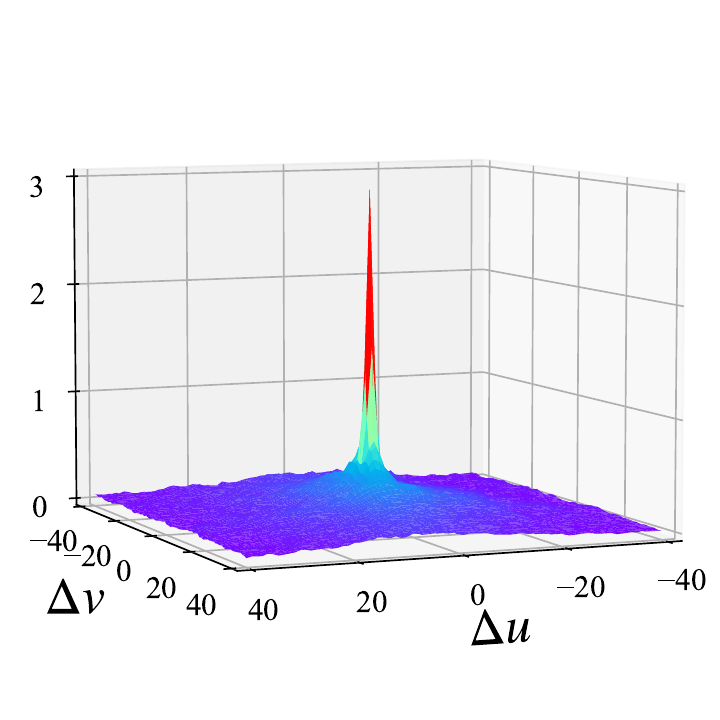}
    \caption{$s10-20$}
  \end{subfigure}
  \begin{subfigure}{0.24\linewidth}
    \includegraphics[width=1\linewidth, trim=0 40 0 40]{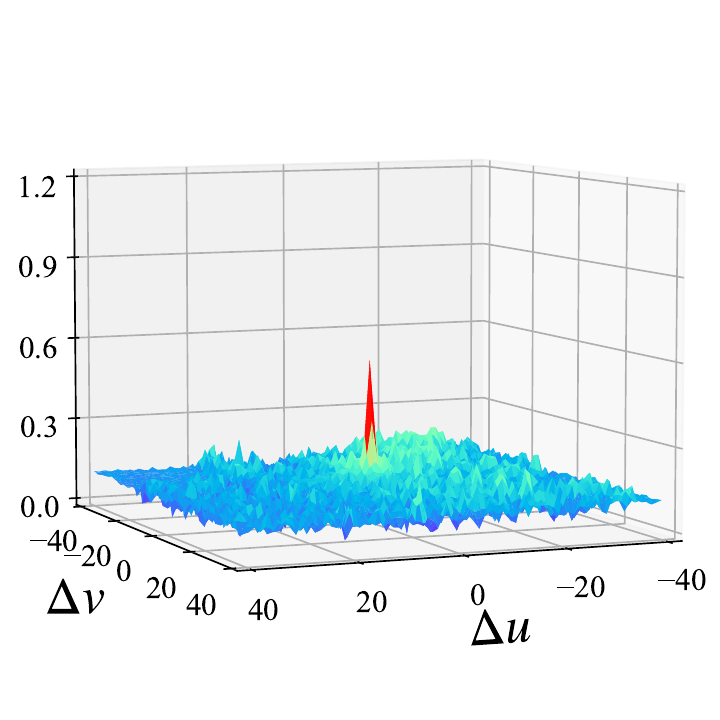}
    \includegraphics[width=1\linewidth, trim=0 40 0 0]{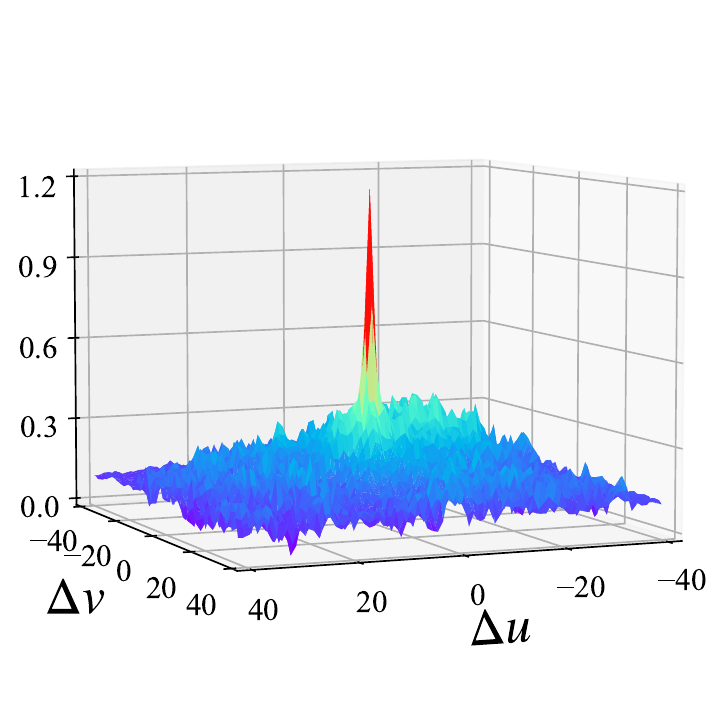}
    \caption{$s20-30$}
  \end{subfigure}
  \begin{subfigure}{0.24\linewidth}
    \includegraphics[width=1\linewidth, trim=0 40 0 40]{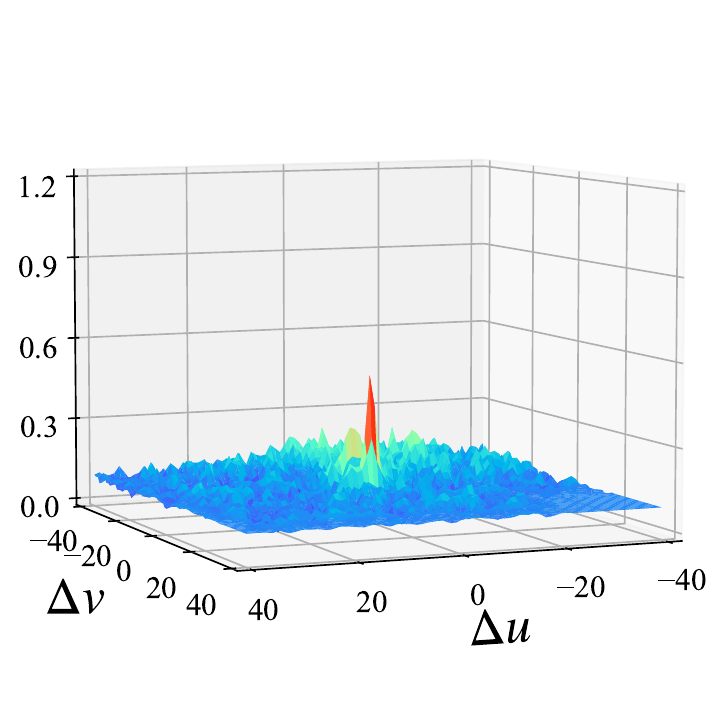}
    \includegraphics[width=1\linewidth, trim=0 40 0 0]{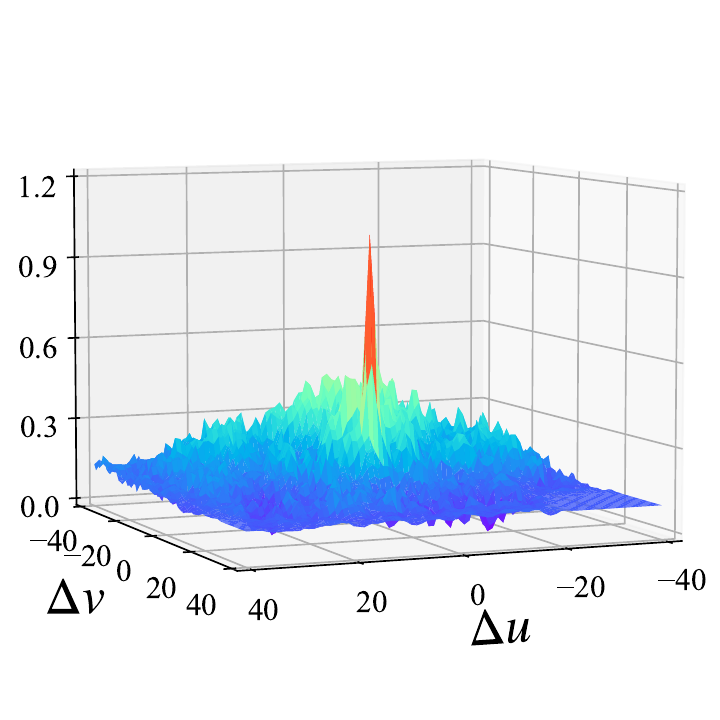}
    \caption{$s30+$}
  \end{subfigure}
  \caption{{\bf Visualization of cost volumes in different range of displacements.} The first row is for RAFT~\cite{teed2020raft}, and the second row is ours. $s10$ refers to regions with displacements below 10 pixels, $s10-20$ for displacements between 10 and 20 pixels, $s20-30$ for displacements between 20 and 30 pixels, and $s30+$ for displacements larger than 30 pixels.} 
  \label{fig:cost_vis}
\end{figure*}

We visualize the cost volume for different ranges of displacements in Fig.~\ref{fig:cost_vis}. 
The larger the value at the center of the averaged matching matrix is, the higher quality the cost volume has.
As shown, GMFlowNet outperforms RAFT in all displacement ranges, which indicates that our approach provides better cost volumes not only for small displacements but also for large ones.

\end{document}